\documentclass[authoryear,3p,preprint,12pt]{elsarticle}
\pdfoutput=1

\usepackage{amsmath}
\usepackage{color}

\usepackage[misc]{ifsym}
\usepackage{subfigure}
\usepackage{array}
\usepackage{epsfig}
\usepackage{enumerate}
\usepackage{multirow}
\usepackage{verbatim}
\usepackage{url}
\usepackage{tabularx}
\usepackage[linesnumbered,ruled,vlined]{algorithm2e}
\usepackage{bbding}
\usepackage{threeparttable}
\usepackage{makecell}
\newcommand{\revise}[1]{{\color{black} #1}}
\usepackage{subcaption}
\usepackage{microtype}
\usepackage{booktabs}
\usepackage{algorithmic}
\usepackage{setspace}
\usepackage{graphicx} 

\usepackage{amssymb}

\usepackage{lineno}

\makeatletter
\def\ps@pprintTitle{%
	\let\@oddhead\@empty
	\let\@evenhead\@empty
	\let\@oddfoot\@empty
	\let\@evenfoot\@oddfoot
}
\makeatother

\begin{document}

\begin{frontmatter}



\title{Low Saturation Confidence Distribution-based Test-Time Adaptation for Cross-Domain Remote Sensing Image Classification}


\author[inst1]{Yu Liang\corref{eq}}
\author[inst1]{Shilei Cao\corref{eq}}
\author[inst1,inst5]{Juepeng Zheng\corref{cor1}}
\author[inst1]{Xiucheng Zhang}
\author[inst0,inst2,inst3]{Jianxi Huang}
\author[inst4,inst5,inst6]{Haohuan Fu}

\affiliation[inst1]{
            organization={School of Artificial Intelligence, Sun Yat-Sen University},
            city={Zhuhai},
            postcode={519080}, 
            country={China}}

\affiliation[inst0]{
            organization={Faculty of Geosciences and Engineering, Southwest Jiaotong University},
            city={Chengdu},
            postcode={611756},
            country={China}
}
            
\affiliation[inst2]{
            organization={College of Land Science and Technology, China Agricultural University},
            city={Beijing},
            postcode={100083}, 
            country={China}}

\affiliation[inst3]{
            organization={Key Laboratory of Remote Sensing for Agri-Hazards, Ministry of Agriculture and Rural Affairs},
            city={Beijing},
            postcode={100083}, 
            country={China}}

\affiliation[inst4]{
            organization={Tsinghua Shenzhen International Graduate School, Tsinghua University},
            city={Shenzhen},
            postcode={518071}, 
            country={China}}
            
\affiliation[inst5]{
            organization={National Supercomputing Center in Shenzhen},
            city={Shenzhen},
            postcode={518055}, 
            country={China}}
            
\affiliation[inst6]{
            organization={Ministry of Education Key Laboratory for Earth System Modeling and the Department of Earth System Science, Tsinghua University},
            city={Beijing},
            postcode={100084}, 
            country={China}}
            
\cortext[eq]{These authors contributed equally to this work.}
\cortext[cor1]{Corresponding author: zhengjp8@mail.sysu.edu.cn}

\begin{abstract}
Unsupervised Domain Adaptation (UDA) has emerged as a powerful technique for addressing the distribution shift across various Remote Sensing (RS) applications.
However, most UDA approaches require access to source data, which may be infeasible due to data privacy or transmission constraints.
Source-free Domain Adaptation addresses the absence of source data but usually demands a large amount of target domain data beforehand, hindering rapid adaptation and restricting their applicability in broader scenarios. 
In practical cross-domain RS image classification, achieving a balance between adaptation speed and accuracy is crucial.
Therefore, we propose Low Saturation Confidence Distribution Test-Time Adaptation (LSCD-TTA), marketing the first attempt to explore Test-Time Adaptation for cross-domain RS image classification without requiring source or target training data. 
LSCD-TTA adapts a source-trained model on the fly using only the target test data encountered during inference, enabling immediate and efficient adaptation while maintaining high accuracy.
Specifically, LSCD-TTA incorporates three optimization strategies tailored to the distribution characteristics of RS images.
Firstly, weak-confidence softmax-entropy loss emphasizes categories that are more difficult to classify to address unbalanced class distribution.
Secondly, balanced-categories softmax-entropy loss softens and balances the predicted probabilities to tackle the category diversity.
Finally, low saturation distribution loss utilizes soft log-likelihood ratios to reduce the impact of low-confidence samples in the later stages of adaptation.
By effectively combining these losses, LSCD-TTA enables rapid and accurate adaptation to the target domain for RS image classification.
We evaluate LSCD-TTA on six domain adaptation tasks across three RS datasets, where LSCD-TTA outperforms existing DA and TTA methods with average accuracy gains of 4.99$\%$ on Resnet-50, 5.22$\%$ on Resnet-101, \revise{and 2.37$\%$ on ViT-B/16}.
\end{abstract}




\end{frontmatter}


\section{Introduction}
\label{sec:intro}
Deep learning has demonstrated remarkable potential in Remote Sensing (RS) image classification tasks, excelling in efficiently and effectively extracting feature information from RS images \citep{li2019learning,teng2019classifier,zheng2024review}.
However, models trained on the source domain often experience significant performance degradation when applied to a target domain with a different distribution—a problem known as distribution shift \citep{geirhos2018generalisation}.
This challenge is compounded in RS imagery due to its diversity in land categories, high information density, and variable meteorological conditions.

\begin{figure}[tbp]
    \centering
    \includegraphics[width=0.7\linewidth]{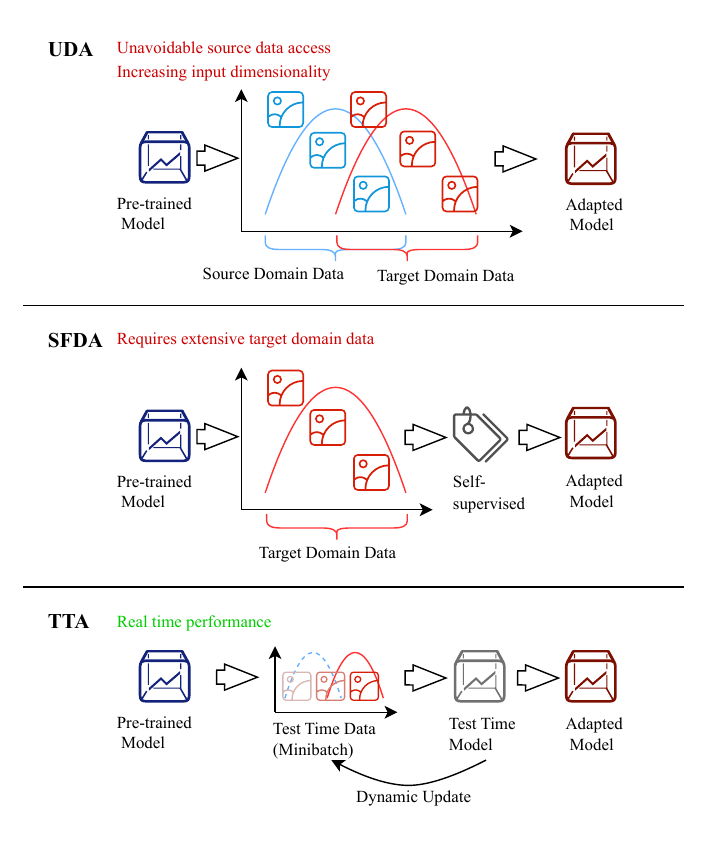}
    \caption{Comparisons of different domain adaptation methods for remote sensing image classification. 
    }
    \label{fig:intro}
\end{figure}

As shown in Fig. \ref{fig:intro}, Unsupervised Domain Adaptation (UDA) has gradually developed to address the distribution shift \revise{between source and target domains by leveraging techniques such as adversarial learning \citep{lee2019drop,zheng2020cross,zheng2024open} and feature alignment \citep{zhang2022spectral,wei2021metaalign,li2024hyunida}.
}
\revise{
Similarly, Domain Generalization (DG) aims to train a model that generalizes well to multiple unseen target domains by learning from multiple source domains \citep{muandet2013domain,zheng2021multisource,liang2024single}.
}
Nevertheless, UDA \revise{and DG} methods are constrained by their dependence on source data.
In many RS scenarios, where data privacy is a concern (e.g., military monitoring and resource detection), data privacy concerns make accessing source data impractical, creating significant obstacles for UDA \revise{and DG} methods.

To overcome this limitation, Source-free Domain Adaptation (SFDA) has been introduced within the RS community \citep{xu2022source,liu2024source}, enabling the adaptation of source-trained models to target training data without requiring access to the source data.
Despite this advancement, typical SFDA methods still adhere to the conventional train-test paradigm, where models are trained on considerable target data and remain fixed during testing.
This process necessitates multiple iterations to progressively align with the target distribution, which is time-consuming and unsuitable for real-world RS applications that demand rapid and robust adaptation.
For example, UAV-based disaster monitoring requires models that can quickly adjust to changing conditions across diverse regions \citep{jiang2022fusion}.
Consequently, traditional Domain Adaptation (DA) models struggle to meet urgent and real-time demands, remaining limited by offline learning constraints. 

Test-Time Adaptation (TTA) has recently emerged as a promising online learning method to directly adapt the source-trained model to target test data during inference \citep{tent}.
\revise{
This process does not require access to source domain datasets or substantial amounts of target data for pre-training, making it particularly suitable for scenarios where immediate adaptation is crucial and data accessibility is limited.
Existing TTA methods concentrate mostly on self-supervised learning methods \citep{slr} and optimization-based strategy \cite{niu2023towards,lame} in the computer vision domain.
For example, 
\citet{lame} employ an efficient concave procedure and Laplace optimization to adjust the maximum likelihood estimation objective, addressing uncertainty during testing.}
Although classic TTA approaches have demonstrated adaptability to changing data, they have primarily been validated on artificially corrupted datasets (e.g., CIFAR10-C \citep{hendrycks2019benchmarking}).
These conditions differ significantly from real-world RS scenarios characterized by multiple cross-domain styles, high information density, variable meteorological environments, and large-scale data \citep{zheng2023surveying}. 
Direct application of existing TTA methods to cross-domain RS image classification requires further optimization to address these unique challenges.
\revise{To date, no work has introduced TTA into RS image classification tasks, particularly for scenarios demanding high-resolution and real-time processing.}

Therefore, we propose Low Saturation Confidence Distribution Test-Time Adaptation (LSCD-TTA) based on the characteristics of RS images.
LSCD-TTA includes three core optimization strategies designed to address the distribution characteristics of RS images, aimed at rapidly and accurately adapting to the target domain.
Specifically, Weak-confidence Softmax-Entropy (WCSE) Loss places greater emphasis on hard-to-distinguish, weak-category RS samples, enhancing the model’s ability to classify these challenging samples with higher accuracy.
Moreover, Balanced-categories Softmax-Entropy (BCSE) Loss softens the confidence levels across categories by refactoring the predicted probability distributions, effectively handling category diversity.
Furthermore, Low Saturation Distribution (LSD) Loss desaturates the traditional cross-entropy loss based on soft log-likelihood ratios.
The LSD loss mitigates the dominance of low-confidence samples during later stages of adaptation, preventing them from disproportionately influencing model updates.
We demonstrate LSCD-TTA's effectiveness on six DA tasks across three RS datasets (\textit{i.e.}, AID \citep{aid}, NWPU-RESISC45 \citep{nwpu}, and UC Merced \citep{umerced}). 
Experimental results show that LSCD-TTA significantly outperforms existing SFDA and TTA methods, achieving average accuracy gains of 4.99\% with ResNet-50, 5.22\% with ResNet-101, \revise{and 2.37$\%$ on ViT-B/16}. 
The method not only improves overall accuracy but enhances robustness, making it well-suited for real-world applications that require rapid adaptation and high precision.

In summary, the main contributions of this paper are as follows:
\begin{enumerate}
    \item
    We propose a TTA method LSCD-TTA for rapid and accurate cross-domain RS image classification.
    To the best of our knowledge, LSCD-TTA is the first attempt to integrate TTA with cross-domain classification tasks within the RS community.
    \item
    LSCD-TTA introduces WCSE, BCSE, and LSD to enhance low-confidence sample performance, improve weak category classification, and mitigate cross-category distribution bias.
    These losses are designed to account for the specific characteristics of RS images, which addresses the challenges when directly applying classic TTA to RS image data. 
    \item
    We conduct extensive experiments and analysis to exhibit the superiority of LSCD-TTA on six DA tasks, achieving remarkable average accuracy improvements over state-of-the-art SFDA and TTA methods of 4.99$\%$ with Resnet-50, 5.22$\%$ with Resnet-101, and \revise{2.37$\%$ with ViT-B/16} in average accuracy.

\end{enumerate}

\section{Methodology}
\label{sec:method}
\begin{figure*}[tbp]
  \centering
    \includegraphics[width=\linewidth]{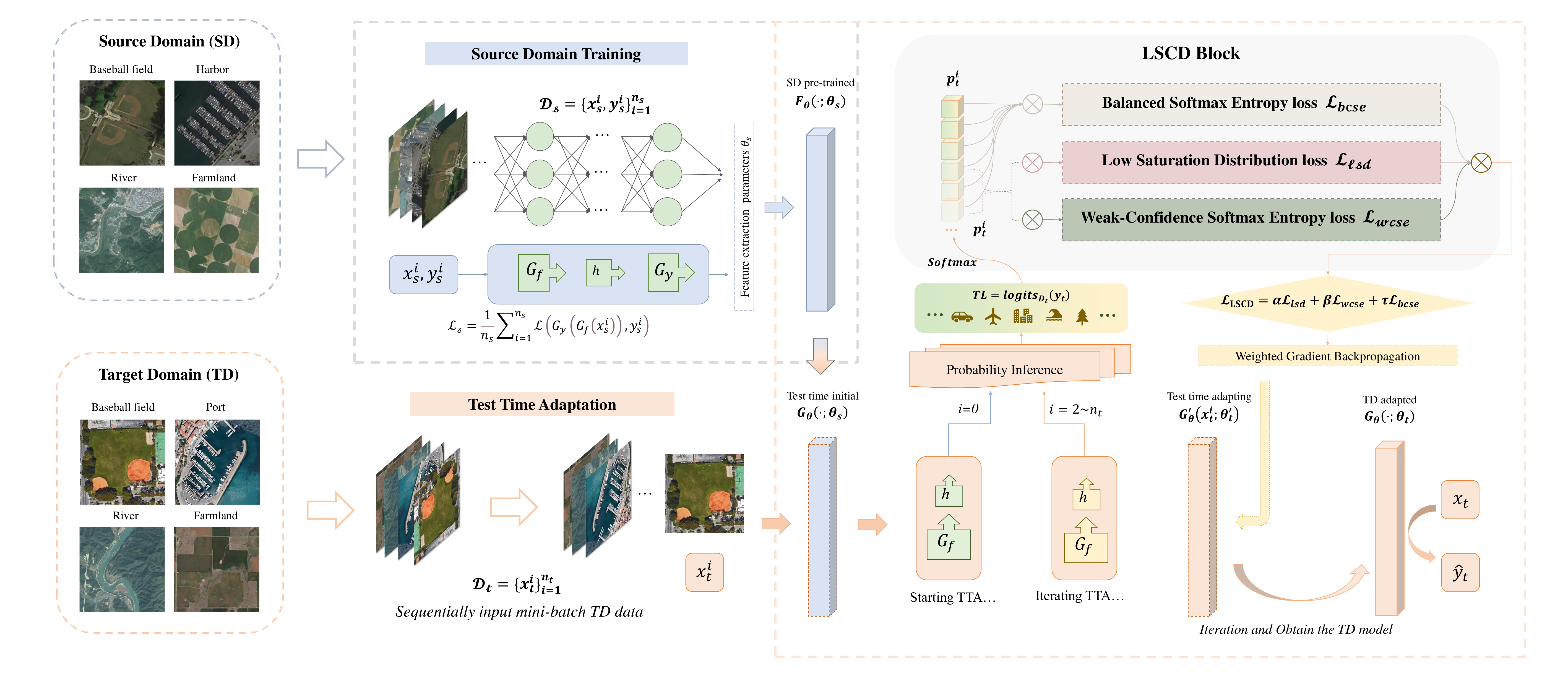}
    \vspace{-3mm}
    \caption{
    The main structure of LSCD-TTA. 
    The model $F_{\theta}$ is first pre-trained on the source domain, which serves as the initialization of $G_{\theta}$ for adaptation toward the target domain. 
    The model extracts the features of target test images within the continuous input and calculates the probability distribution using model $G_{\theta}$ during adaptation.
    Subsequently, LSCD-TTA balances the predicted probabilities across categories (\textit{i.e.} $\mathcal{L}_{bcse}$) and especially emphasizes low-saturation or week-probability samples that are difficult to distinguish (\textit{i.e.} $\mathcal{L}_{lsd}$ and $\mathcal{L}_{wcse}$), to align the source and target domains in real time.
    }
    \label{fig:main}
\end{figure*}

In this paper, we propose Low Saturation Confidence Distribution Test-Time Adaptation (LSCD-TTA), which is designed for cross-scene RS image classification tasks.
LSCD-TTA includes three losses that consider weak-category samples, category diversity, and low saturation respectively (See Fig. \ref{fig:main}).
By integrating these losses, LSCD-TTA significantly enhances the model’s generalization capability to the target domain, effectively handling the complexity inherent in cross-scene RS image classification. 

\subsection{Problem preliminaries}
Let $\mathcal{D}s = \{x_s^i,y_s^i\}_{i=1}^{n_s}$ denote the source domain dataset, where $x_s^i$ represents the input images and $y_s^i$ the corresponding labels. 
Similarly, let $\mathcal{D}t = \{x_t^i\}_{i=1}^{n_t}$ denote the target domain dataset, consisting of unlabeled images $x_t^i \in \mathcal{X}_t$. 
Our goal is to adapt a model $G_\theta: \mathcal{X} \rightarrow \mathcal{Y}$, initialized by source-trained model $F_\theta$, to perform well on the target domain without accessing source data. 
We aim to adjust the model’s parameters $\theta$ during test time using only the unlabeled target data $\mathcal{D}_t$, enabling the model to generate an accurate mapping $G: \mathcal{X}_t \rightarrow \mathcal{Y}_t$.
Specifically, given target test data $x_t^i$ at batch i, the model is first to make a prediction ${y}_t^i = G(x_t^i)$ using the parameters $\theta^{i-1}$ which have been updated based on previous target data $x_t^1, ..., x_t^{i-1}$. 
Subsequently, ${y}_t^i$ serves as the evaluation output at batch i, and the model will adapt itself toward $x_t^i$ as $\theta^{i-1} \rightarrow \theta^{i}$, which will only influence future inputs $x_t^{i+n}$.

\subsection{Parameter iteration}
Motivated by Tent \citep{tent}, the foundation work of TTA, our approach exclusively updates transformation parameters $\gamma, \revise{b}$ in batchnorm layers through backpropagation to enable robust and efficient adaptation. 
Moreover, \revise{t}he normalized statistics $\mu,\sigma$ in batchnorm layers during the testing process are re-estimated based on target test data:
\begin{equation}
\begin{aligned}
    & \mu_{t}=\frac{1}{n_t}\sum_{i=1}^{n_t}x_{t}^{i}\\
    & \sigma_{t}^{2}=\frac{1}{n_t}\sum_{i=1}^{n_t}(x_{t}^{i}-\mu_{t})^{2}
\end{aligned}
\end{equation}
Subsequently, for a given input batch $x_t$, a normalization operation is performed on the data:
\begin{equation}
    {x_t}=\frac{x_t-\mu_{t}}{\sqrt{\sigma_{t}^{2}+\epsilon}}
\end{equation}
In the optimization transformation process of the backpropagation stage, the transformation parameters include the scaling parameter $\gamma$ and the bias parameter $\revise{b}$ of the Batch Normalization (BN) layer.
For each input $x_t$ in the target domain, the output of the BN layer can be denoted as:
\begin{equation}
    y=\gamma{x_t}+\revise{b}
\end{equation}
In general, one can compute the gradient of these parameters $\frac{\partial\mathcal{L}}{\partial\gamma},\frac{\partial\mathcal{L}}{\partial\revise{b}}$ by means of the loss function $\mathcal{L}$ at the time of testing process, and subsequently back-propagate the gradient to perform the optimization of the transformed parameters:
\begin{equation}
\begin{aligned}
    & \gamma = \gamma + \frac{\partial\mathcal{L}}{\partial\gamma} \\
    & \revise{b} = \revise{b} + \frac{\partial\mathcal{L}}{\partial\revise{b}}
\end{aligned}
\end{equation}

        


        



The normalized statistics of the target domain data and the transformation parameters are iteratively updated at each step of the adaptation process. 
Notably, as mentioned in Section \ref{sec:intro}, TTA is designed within an online learning framework, allowing iterative updates as long as the target domain data stream remains continuous.

\subsection{Weak-confidence Softmax-Entropy loss $\mathcal{L}_{\revise{wcse}}$}
In classical supervised classification tasks, cross-entropy loss is commonly used to measure the discrepancy between a model’s predicted probability distribution and the ground-truth labels.
Let ${y}^g$ denote the one-hot ground-truth labels and ${y}$ represent the predicted probability distribution. The cross-entropy loss $H$ is formulated as:
\begin{equation}
    \label{eq:h_normal}
    \begin{aligned}
      H({y},{y}^g) &= -\sum_{c=1}^C{y}_c^g\log {y}_c
    \end{aligned}
\end{equation}
where $C$ denotes the number of categories.


In TTA scenarios, the ground-truth labels for the target data are not available. 
A straightforward strategy is to adopt hard pseudo-label self-supervision by converting the predicted probability distribution into one-hot hard pseudo-labels, thus assuming complete trust in the model’s predictions. 
The corresponding hard pseudo-label cross-entropy loss is:
\begin{equation}
    \label{eq:h_hard}
    \begin{aligned}
    {y}_{hard} &= 
    \begin{cases}
    1  & \text{if } c = \hat{c} \\
    0 & \text{if } c \neq \hat{c}
    \end{cases} \\
     H({y}, {y}^{hard}) &= -\sum_{c=1}^C{y}_c^{hard}\log {y}_c = -y_{\hat{c}} \log y_{\hat{c}}\\
    \end{aligned}
\end{equation}
where $\hat{c} = \underset{c}{\operatorname{argmax}}({y_c})$ corresponding to
the class with maximal confidence. 
While this hard pseudo-labeling framework is simple and often effective, it disproportionately emphasizes the categories that the model is already confident about.
This is problematic when the target domain exhibits substantial shifts from the source domain, especially in RS imagery where certain classes are inherently difficult to distinguish due to high inter-class similarity. 
Consequently, the model’s gradient updates might neglect critical information from these hard-to-classify samples, resulting in suboptimal adaptation.

The soft pseudo-label approach introduced in Tent (\textit{i.e.}, directly utilizes the model prediction ${y}$ as the pseudo-labels) takes uncertainty into account by leveraging cross-entropy loss \citep{tent}.
Although such a method is widely adopted for many TTA tasks, it remains insufficient for RS imagery, which often contains complex features and subtle inter-class differences. 
Moreover, no existing methods fully leverage weak category information from unlabeled target data. To address these issues, we propose the Weak-Confidence Softmax-Entropy (WCSE) loss, which aims to enhance the learning of weak categories and improve classification accuracy in TTA for RS imagery.

Specifically, we propose a weak-category distribution density ${d}$ designed to highlight categories that are less likely or harder to classify:
\begin{equation}
    \label{eq:soft_one_hot}
    {d}_{week} = 
    \begin{cases}
    \epsilon  & \text{if } c = \hat{c} \\
    1 - \frac{\epsilon}{C-1} & \text{if } c \neq \hat{c}
    \end{cases}
\end{equation}
where $\epsilon$ is a smoothing hyper-parameter and \revise{is} recommended to be set to 0.01.

Subsequently, we employ an exponential function to scale ${d}$, assigning the week-category confidence weight to the categories:
\begin{equation}
    \label{eq:wcse_1}
    \begin{aligned}
    {\lambda}_{wc} &= \exp({{d}_{week}})  \\
    \end{aligned}
\end{equation}
This exponential scaling ensures the importance of probabilities is appropriately re-emphasized to the desired scale.

Finally, we utilize the square root of the prediction as the pseudo-label to form smoother pseudo-labels. 
Such adjustment enhances the model's adaptability and robustness to multi-class distributions in RS imagery. 
Utilize the week-category confidence weight, the WCSE loss is defined as:
\begin{equation}
    \label{eq:wcse}
     \mathcal{L}_{\revise{wcse}} = H({y}, {\lambda}_{wc} \odot \sqrt{{y}}) = -\sum_{c=1}^C {\lambda}_{c}\sqrt{{y}_c}\log {y}_c  
\end{equation}

where $\odot$ denotes the element-wise multiplication.
By assigning higher weights to poorly predicted or underrepresented categories, WCSE encourages the model to pay more attention to weak-category samples, thereby improving overall classification performance.

\subsection{Balanced-categories Softmax-Entropy loss $\mathcal{L}_{\revise{bcse}}$}
In conventional DA paradigms, it is often presumed that distributional discrepancies between source and target domains are predominantly confined to the data boundaries. 
However, this assumption does not hold uniformly across all applications. 
Specifically, in RS image classification, domain shifts frequently manifest not only at the distribution peripheries but also centrally within the data distribution. 
For instance, a particular land cover class may exhibit diverse spectral signatures under varying environmental conditions or across different geographic regions, leading to intra-class variability that complicates accurate classification. 
This central distributional shift is exacerbated by inherent class imbalances, where certain categories are underrepresented, thereby increasing the propensity for misclassification.

This phenomenon, which arises from class imbalance rather than edge noise, is particularly prevalent in RS scenarios. 
Such discrepancies can lead to classification errors or a decline in model performance when adapting to the target domain.
Accordingly, this characteristic of RS images calls for a method capable of handling the categories distribution across the entire region in environment-varying scenarios while maintaining robustness in real-time cross-domain adaptation.

To address the challenge of categorical diversity, we propose the Balanced-Categories Softmax-Entropy (BCSE) loss.
This loss function is designed to soften and balance the predicted probability distributions, thereby enhancing the model’s ability to generalize across diverse categories in the target domain.
Specifically, we introduce a balanced distribution representation ${b}$ that integrates the weak-category distribution density $d_{week}$ defined earlier.
Furthermore, we still scale ${b}$ using the exponential function, thereby further enhancing robustness. 
Such a design can be expressed as:
\begin{equation}
    \label{eq:bcse_1}
    \begin{aligned}
      {b} &= {y} \odot (1-{d}_{weak}) + (1-{y}) \odot {d}_{weak} \\
    \lambda_{bc} &= \exp({{b}}) 
    \end{aligned}
\end{equation}
Subsequently, the BCSE loss is formulated by weighting the standard softmax-entropy loss with the balanced category coefficients  $\lambda_{bc}$:
\begin{equation}
    \label{eq:bcse}
     \mathcal{L}_{\revise{bcse}}  = H(y, \lambda_{bc}\odot{y}) = - \sum_{c=1}^C\lambda_{c}{y}_c\log {y}_c    
\end{equation}
The BCSE loss function appropriately adjusts the contribution of each class to the overall loss based on its representation and confidence level. 
Specifically, classes that are underrepresented or less confidently predicted receive higher weights, thereby encouraging the model to allocate more learning capacity to these categories. This balanced weighting mitigates the risk of the model becoming biased towards dominant classes, promoting a more equitable adaptation across all categories in the target domain. Consequently, the BCSE loss enhances the model’s ability to generalize effectively in environments with significant categorical diversity and imbalance, which are characteristic of RS imagery.

\subsection{Low Saturation Distribution loss $\mathcal{L}_{\revise{lsd}}$}
Traditional entropy-based loss functions, such as those employed in methods like TENT \citep{tent}, tend to suffer from vanishing gradients when the model’s predictions approach near-certain values (i.e., probabilities near one). 
Consequently, as the adaptation progresses and predictions become more confident, the gradient signals derived from high-confidence samples diminish, leading to a dominance of low-confidence samples in the loss computation. 
This imbalance is particularly problematic in RS scenarios, where inter-class similarities and environmental variations can make certain categories inherently difficult to classify. 
Therefore, there is a pressing need for a loss function that mitigates the overconfidence in high-confidence categories while preserving informative gradient signals from all samples, irrespective of their confidence levels.

To address the issue of distributional saturation, we introduce the Low Saturation Distribution (LSD) loss, which is designed to regulate the influence of high-confidence predictions and maintain sensitivity to low-confidence samples during the adaptation process. The LSD loss builds upon the concept of likelihood ratio penalties, which are adept at preserving non-vanishing gradients even when the model exhibits high confidence in its predictions.
Inspired by the limitations of both hard and soft pseudo-labeling approaches, the LSD loss employs a ratio-based penalty term that inversely relates to the collective probability mass of all classes excluding the predicted class. Formally, the LSD loss is defined as:
\begin{equation}
    \label{eq:lsd1}
   \mathcal{L}_{\revise{lsd}} = -\sum_{c=1}^C{y}_c\log\frac{1}{\sum_{i\neq {c}}{y}_i} = \sum_{i=c}^C{y}_c(\log\sum_{i\neq {c}}{y}_i)
\end{equation}

Specifically, the term $\log\sum_{i\neq {c}}{y}_i$ ensures that as the confidence in class c increases (i.e., $y_c \rightarrow$ 1), the sum $\sum_{i \neq c} y_i$ decreases, thereby preventing the logarithmic term from vanishing. 
This design choice guarantees that the gradient signals remain substantial even for high-confidence predictions, thereby avoiding the dominance of low-confidence samples in the loss function.
By integrating the LSD loss into the overall adaptation objective, we effectively reduce the model’s overreliance on dominant classes. This encourages the model to allocate more learning capacity to weakly represented or ambiguous samples, which are critical for capturing subtle domain shifts and enhancing classification robustness in RS imagery. 



Finally, combining these three modules together, the resulting composite loss, $\mathcal{L}_{\revise{lscd}}$, is:
\begin{equation}
    \label{eq:LSCD}
    \mathcal{L}_{\revise{lscd}}= \alpha\mathcal{L}_{\revise{wcse}} + \beta\mathcal{L}_{\revise{bcse}} + \tau\mathcal{L}_{\revise{lsd}}
\end{equation}
where $\alpha$, $\beta$, $\tau$ are hyperparameters controlling the balance among these three modules. 
This composite objective is readily applied to an online or offline TTA procedure, offering a fully test-time adaptation framework that addresses high-dimensional, cross-scene RS datasets more effectively than conventional soft or hard pseudo-labeling strategies.

\begin{figure}[tbp]
    \centering
    \includegraphics[width=\linewidth]{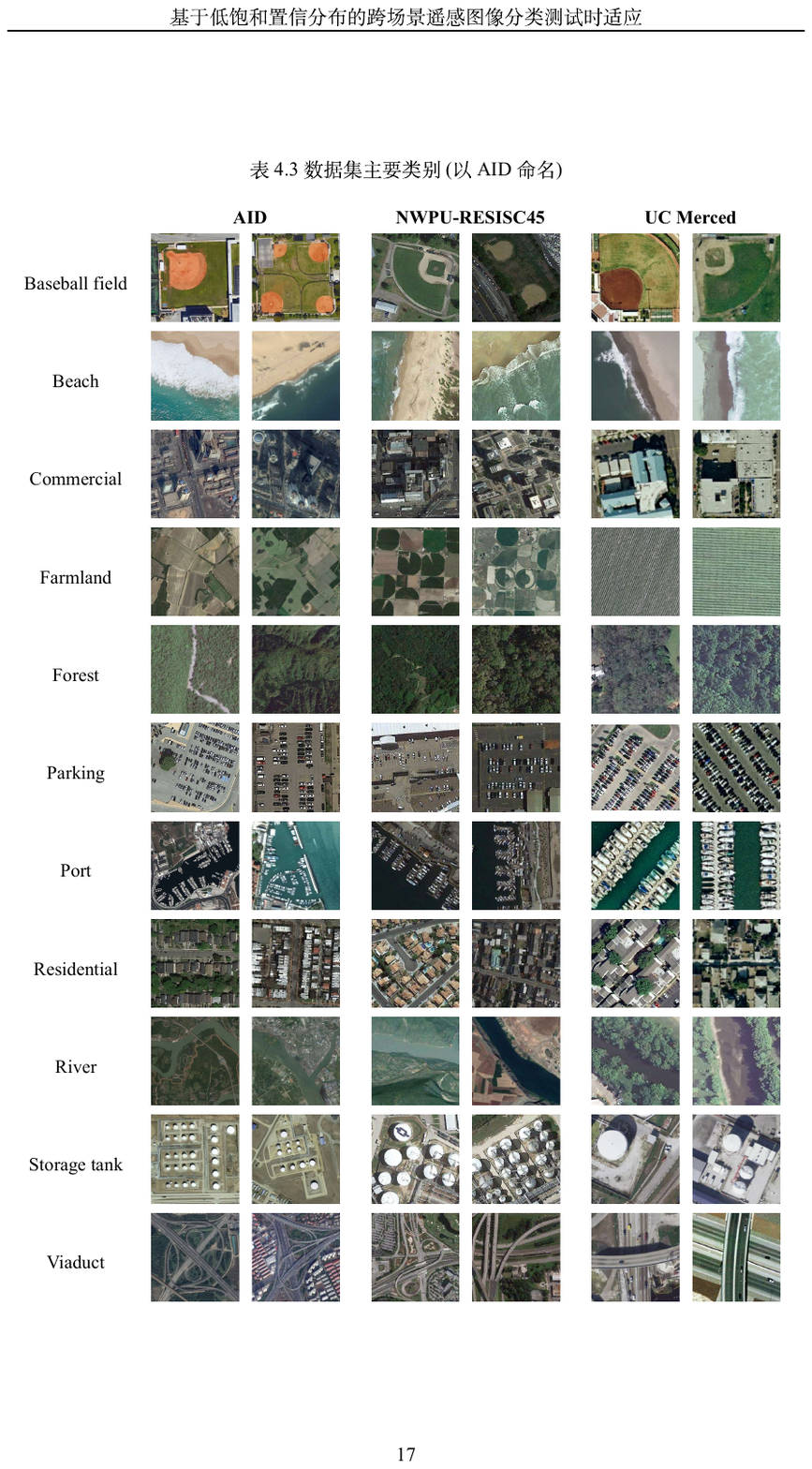}
    \caption{Samples of the datasets}
    \label{fig:expall}
\end{figure}

\section{Datasets}
\label{sec:dataset}
As shown in Fig.\ref{fig:expall}, the RS data used for the experiments in this study are based on three different open-source RS datasets: AID \citep{aid}, NWPU-RESISC45 \citep{nwpu} and UC Merced \citep{umerced}:
\begin{enumerate}
    \item \textbf{NWPU-RESISC45}
        contains high-resolution aerial images derived from 45 scene classes, each with 700 images of size 256 × 256. 
        NWPU-RESISC45 is mainly used for tasks such as target recognition, classification, and scene understanding, such as buildings, forests, and lakes.
    \item \textbf{AID}
        contains 10 cities from Google Earth and other platforms, and consists of panoramic RS images covering 30 land use categories, which cover a wide range of conditions in different seasons and weather.
        AID is a standard dataset for evaluating the performance of aerial scene classification algorithms because of its rich scene categories and scale with 10000 images in total. 
    \item \textbf{UC Merced}
        consists of aerial imagery taken by a digital camera mounted on a civilian airplane, including 21 land use categories.
        UC Merced provides high-resolution imagery with a spatial resolution of 0.3 meters per pixel. It is commonly used for land use classification, scene understanding, and urban planning, totaling 2100 images.
\end{enumerate}
We conduct six DA experiments by pairing these three datasets in all possible combinations.
The experiments select a shared class between two domains to ensure that the source and target domains have consistent classes following previous studies \citep{zheng2021two,zheng2022partial,guo2024c,zheng2024open,chen2025ban}.
Note that the class names of corresponding categories may differ between the two datasets.
For example, in the transfer task NWPU-RESISC45$\rightarrow$AID, both the circular farmland and rectangular farmland classes in NWPU-RESISC45 correspond to the farmland classes in AID.

\section{Experiments}
\label{sec:exp}
\subsection{Experiments Setup}
In this paper, six DA tasks are constructed to evaluate the models' real-time adaptation performance to unseen domains with significant distributional gaps, leveraging only the source model.
The six DA tasks are defined as follows: A$\rightarrow$N, N$\rightarrow$A; A$\rightarrow$U, U$\rightarrow$A; U$\rightarrow$N, and N$\rightarrow$U, where A, U, and N represent the AID, UC Merced, and NWPU-RESISC45 datasets, respectively. 
NWPU-RESISC45 and AID have the largest overlap in shared categories for DA, with 23 shared classes. In contrast, UC Merced and AID share only 13 categories. 
Data augmentation techniques, such as rotation and cropping, are applied to improve the generalization capability during data preprocessing.
Additionally, all experiments utilize stochastic gradient descent with an initial learning rate of 0.001 and a momentum of 0.9 as the optimizer for the adapted model during the TTA process.
\revise{Moreover, the experiments in this study are developed on PyTorch deep learning framework. All computations are performed on an NVIDIA A800 GPU. During model training, we adopt an 80/20 data split strategy. To further enhance the generalization ability of the model, we employ cross-validation, which effectively reduces the risk of overfitting. For the activation function, we select ReLU, which has demonstrated excellent performance in most computer vision tasks.}

\subsection{Baseline Methods}
We compare LSCD-TTA against various state-of-the-art DA and TTA methods, utilizing ResNet-50, ResNet-101 \citep{resnet}, \revise{and ViT-B/16 \citep{dosovitskiy2020image}} employed as backbones.
The comparison includes the source-trained model as the ``Baseline'', as well as methods employing confidence-based loss, cross-entropy (CE) loss, and KL divergence \citep{KL}, which serve as classical domain alignment techniques based on the pseudo label. 
Furthermore, other DA methods such as CBST \citep{cbst}, ADDA \citep{adda}, \revise{ARS \citep{ARS}, CS-CADA \citep{CS-CADA},} and SHOT \citep{shot} are also included, which aim to minimize the distributional gap between source and target domains through self-training, adversarial training, or the transfer of source-domain hypothesis information.
Additionally, LSCD-TTA is compared with existing TTA methods, including Tent \citep{tent}, SLR \citep{slr}, LAME \citep{lame}, ARM \citep{arm}, DomainAdaptator \citep{gem}, \revise{and TRIBE \citep{TRIBE}}.

\begin{table}[htbp]
\centering
\centering
\caption{Accuracy and standard deviation (\%) of comparative study on RS datasets (Backbone: \textit{Resnet-50})}
\label{tab:methods_50}
\renewcommand\arraystretch{1.0}
\resizebox{\linewidth}{!}{
\begin{tabular}{p{2.8cm}<{\centering}|cc|cc|cc|c}
\hline
\multirow{2}{*}{\textbf{Method}} & \multicolumn{2}{|c|}{NWPU-RESISC45} & \multicolumn{2}{|c|}{AID} & \multicolumn{2}{|c|}{UC Merced} & \multirow{2}{*}{\begin{tabular}{c}
    \footnotesize  \textbf{Average}\\
    \footnotesize  \textbf{Accuracy}
\end{tabular}}  \\ 
\cline{2-7}
                         & \textbf{N$\rightarrow$A}      & \textbf{N$\rightarrow$U}     & \textbf{A$\rightarrow$N}        & \textbf{A$\rightarrow$U}        &\textbf{U$\rightarrow$A}        & \textbf{U$\rightarrow$N}              &                        \\
                         \hline
Baseline                & $89.74_{\pm0.58}$  &  $83.98_{\pm1.36}$  &  $78.77_{\pm0.62}$  &  $72.49_{\pm0.67}$  &  $62.11_{\pm0.62}$  &  $58.04_{\pm0.64}$	 & 74.19\\
Confidence              & $88.50_{\pm0.28}$  &  $83.92_{\pm1.18}$  &  $73.72_{\pm1.14}$  &  $72.88_{\pm1.27}$  &  $57.33_{\pm0.82}$  &  $53.13_{\pm0.63}$	 & 71.58\\
CE                      & $89.39_{\pm0.30}$  &  $84.75_{\pm1.26}$  &  $78.81_{\pm0.55}$  &  $74.48_{\pm1.28}$  &  $62.48_{\pm1.01}$  &  $61.49_{\pm1.34}$	 & 75.23\\
KL                      & $89.37_{\pm0.30}$  &  $84.75_{\pm1.26}$  &  $78.83_{\pm0.58}$  &  $74.48_{\pm1.30}$  &  $62.43_{\pm0.86}$  &  $61.63_{\pm1.19}$	 & 75.25\\
CBST                    & $89.19_{\pm0.28}$  &  $85.10_{\pm1.55}$  &  $78.09_{\pm0.55}$  &  $74.17_{\pm1.96}$  &  $62.91_{\pm1.19}$  &  $60.29_{\pm0.93}$	 & 74.96\\
SHOT                    & $89.60_{\pm0.24}$  &  $85.66_{\pm1.19}$  &  $79.56_{\pm0.54}$  &  $75.29_{\pm1.44}$  &  $65.67_{\pm0.62}$  &  $64.00_{\pm0.52}$	 & 76.63\\
ADDA                    & $88.66_{\pm0.22}$  &  $84.56_{\pm1.29}$  &  $75.48_{\pm0.72}$  &  $73.46_{\pm1.23}$  &  $58.45_{\pm0.65}$  &  $57.07_{\pm0.46}$	 & 72.95\\
AdaBN                   & $89.52_{\pm0.18}$  &  $84.92_{\pm1.51}$  &  $78.89_{\pm0.49}$  &  $74.58_{\pm1.33}$  &  $63.00_{\pm0.84}$  &  $61.79_{\pm1.01}$	 & 75.45\\
{ARS}              & $89.30_{\pm0.25}$  &  
$85.28_{\pm1.21}$  & 
$78.84_{\pm0.69}$  &  $74.28_{\pm1.35}$  &  $62.43_{\pm1.01}$  &  $61.20_{\pm0.59}$  & 
{75.22}
\\
{CS-CADA}          & $89.39_{\pm0.28}$  &  $84.79_{\pm1.29}$  &  $79.01_{\pm0.57}$  &  $74.52_{\pm1.25}$  &  $62.59_{\pm0.86}$  &  $61.61_{\pm1.37}$  & 
{75.32}
\\
ARM                     & $89.03_{\pm0.22}$  &  $84.31_{\pm0.50}$  &  $78.20_{\pm0.27}$  &  $72.66_{\pm0.93}$  &  $59.59_{\pm0.47}$  &  $58.48_{\pm0.38}$	 & 73.71\\
LAME                    & $89.02_{\pm0.23}$  &  $84.57_{\pm1.30}$  &  $77.25_{\pm0.53}$  &  $73.60_{\pm1.28}$  &  $59.87_{\pm0.83}$  &  $58.78_{\pm0.49}$	 & 73.85\\
Tent                    & $89.81_{\pm0.17}$  &  $85.30_{\pm1.48}$  &  $79.32_{\pm0.50}$  &  $75.05_{\pm1.26}$  &  $64.32_{\pm0.78}$  &  $62.70_{\pm1.13}$	 & 76.08\\
SLR                     & $91.76_{\pm0.74}$  &  $87.24_{\pm2.58}$  &  $71.93_{\pm3.73}$  &  $79.06_{\pm1.30}$  &  $69.86_{\pm0.85}$  &  $55.85_{\pm2.40}$	 & 75.95\\
DA-T                    & $90.34_{\pm0.56}$  &  $85.31_{\pm1.49}$  &  $79.17_{\pm0.56}$  &  $74.42_{\pm0.63}$  &  $63.27_{\pm0.80}$  &  $59.34_{\pm0.70}$	 & 75.31\\
DA-SKD                  & $90.72_{\pm0.52}$  &  $86.03_{\pm1.55}$  &  $79.40_{\pm0.60}$  &  $75.37_{\pm0.72}$  &  $64.22_{\pm0.89}$  &  $59.69_{\pm0.74}$	 & 75.91\\
DA-AUG                  & $91.05_{\pm0.45}$  &  $85.69_{\pm1.74}$  &  $79.48_{\pm0.59}$  &  $75.09_{\pm0.33}$  &  $65.39_{\pm1.03}$  &  $60.48_{\pm0.76}$	 & 76.2\\
{TRIBE} & {$90.31_{\pm0.39}$} & {$86.03_{\pm1.53}$} & {$80.07_{\pm0.45}$} & {$76.02_{\pm0.60}$} & {$63.13_{\pm1.20}$} & {$60.38_{\pm0.55}$} & {75.99}\\
\hline
\textbf{LSCD-TTA}      & $\textbf{92.32}_{\pm0.25}$  &  $\textbf{88.73}_{\pm1.31}$  &  $\textbf{81.79}_{\pm1.31}$  &  $\textbf{81.46}_{\pm1.47}$  &  $\textbf{76.36}_{\pm1.01}$  &  $\textbf{69.07}_{\pm2.22}$	 & \textbf{81.62}\\
\hline
\end{tabular}
}
\end{table}
    
\begin{table}[htbp]
\centering
\caption{Accuracy and standard deviation (\%) of comparative study on RS datasets (Backbone: \textit{Resnet-101})}
\label{tab:methods_101}
\renewcommand\arraystretch{1.0}
\resizebox{\linewidth}{!}{
\begin{tabular}{p{2.8cm}<{\centering}|cc|cc|cc|c}
\hline
\multirow{2}{*}{\textbf{Method}} & \multicolumn{2}{|c|}{NWPU-RESISC45} & \multicolumn{2}{|c|}{AID} & \multicolumn{2}{|c|}{UC Merced} & \multirow{2}{*}{\begin{tabular}{c}
    \footnotesize  \textbf{Average}\\
    \footnotesize  \textbf{Accuracy}
\end{tabular}}  \\ 
\cline{2-7}
                         & \textbf{N$\rightarrow$A}      & \textbf{N$\rightarrow$U}     & \textbf{A$\rightarrow$N}        & \textbf{A$\rightarrow$U}        &\textbf{U$\rightarrow$A}        & \textbf{U$\rightarrow$N}              &                        \\
                         \hline
Baseline                & $89.98_{\pm0.49}$  &  $87.26_{\pm0.69}$  &  $80.08_{\pm0.70}$  &  $71.81_{\pm0.52}$  &  $63.42_{\pm2.76}$  &  $58.60_{\pm0.39}$	 & 75.19\\
Confidence              & $88.81_{\pm0.46}$  &  $85.87_{\pm0.48}$  &  $73.34_{\pm1.48}$  &  $73.31_{\pm1.03}$  &  $57.02_{\pm2.75}$  &  $48.37_{\pm1.42}$	 & 71.12\\
CE                      & $89.79_{\pm0.37}$  &  $87.10_{\pm0.41}$  &  $79.22_{\pm1.00}$  &  $75.34_{\pm1.10}$  &  $64.63_{\pm2.56}$  &  $60.35_{\pm0.65}$	 & 76.07\\
KL                      & $89.77_{\pm0.34}$  &  $87.10_{\pm0.41}$  &  $79.34_{\pm1.00}$  &  $75.34_{\pm1.10}$  &  $64.62_{\pm2.51}$  &  $60.29_{\pm0.35}$	 & 76.08\\
CBST                    & $89.53_{\pm0.46}$  &  $86.99_{\pm0.59}$  &  $78.13_{\pm0.93}$  &  $74.94_{\pm1.20}$  &  $65.27_{\pm2.46}$  &  $59.02_{\pm0.47}$	 & 75.65\\
SHOT                    & $90.10_{\pm0.36}$  &  $87.46_{\pm0.53}$  &  $80.58_{\pm0.58}$  &  $76.08_{\pm0.92}$  &  $67.77_{\pm2.46}$  &  $63.98_{\pm0.29}$	 & 77.67\\
ADDA                    & $89.17_{\pm0.45}$  &  $86.42_{\pm0.51}$  &  $75.04_{\pm1.50}$  &  $73.85_{\pm0.99}$  &  $59.35_{\pm2.85}$  &  $55.68_{\pm0.81}$	 & 73.25\\
AdaBN                   & $90.02_{\pm0.28}$  &  $87.28_{\pm0.59}$  &  $79.92_{\pm0.80}$  &  $75.63_{\pm0.62}$  &  $65.09_{\pm2.66}$  &  $61.35_{\pm0.49}$	 & 76.55\\
{ARS}                     & $89.79_{\pm0.29}$  &  $87.08_{\pm0.50}$  &  $79.35_{\pm0.66}$  &  $75.23_{\pm0.90}$  &  $64.17_{\pm2.54}$  &  $60.98_{\pm0.31}$  & 
{76.10}\\
{CS-CADA}          & $89.84_{\pm0.34}$  &  $87.15_{\pm0.48}$  &  $79.48_{\pm0.67}$  &  $75.48_{\pm1.06}$  &  $64.76_{\pm2.63}$  &  $60.56_{\pm0.35}$  & 
{76.21}\\
ARM                      & $89.65_{\pm0.27}$  &  $86.78_{\pm0.33}$  &  $78.76_{\pm0.33}$  &  $73.98_{\pm0.73}$  &  $61.83_{\pm1.42}$  &  $58.58_{\pm0.19}$	 & 74.93\\
LAME                     & $90.16_{\pm0.35}$  &  $87.29_{\pm0.54}$  &  $78.34_{\pm0.90}$  &  $76.31_{\pm0.42}$  &  $61.72_{\pm3.11}$  &  $58.06_{\pm0.38}$	 & 75.31\\
Tent                     & $90.39_{\pm0.27}$  &  $87.64_{\pm0.59}$  &  $80.66_{\pm0.85}$  &  $76.20_{\pm0.66}$  &  $66.71_{\pm2.59}$  &  $62.18_{\pm0.68}$	 & 77.30\\
SLR                      & $92.37_{\pm1.05}$  &  $89.40_{\pm0.93}$  &  $73.39_{\pm2.49}$  &  $79.74_{\pm1.01}$  &  $70.73_{\pm2.82}$  &  $44.63_{\pm3.08}$	 & 75.04\\
DA-T                     & $90.15_{\pm0.36}$  &  $87.78_{\pm0.72}$  &  $78.92_{\pm0.76}$  &  $76.54_{\pm0.99}$  &  $63.60_{\pm2.68}$  &  $59.75_{\pm0.38}$	 & 76.12\\
DA-SKD                   & $90.44_{\pm0.33}$  &  $88.66_{\pm0.69}$  &  $79.30_{\pm0.79}$  &  $78.17_{\pm0.71}$  &  $65.11_{\pm2.62}$  &  $60.40_{\pm0.28}$	 & 77.01\\
DA-AUG                   & $90.82_{\pm0.32}$  &  $88.70_{\pm0.51}$  &  $79.64_{\pm0.79}$  &  $77.74_{\pm0.49}$  &  $66.24_{\pm2.48}$  &  $61.25_{\pm0.25}$	 & 77.40\\
{TRIBE} & {$90.69_{\pm0.40}$} & {$88.75_{\pm0.42}$} & {$80.53_{\pm0.72}$} & {$76.86_{\pm0.99}$} & {$64.66_{\pm2.63}$} & {$61.07_{\pm0.59}$} & {77.09}\\

\hline
\textbf{LSCD-TTA}       & $\textbf{92.43}_{\pm0.44}$  &  $\textbf{90.65}_{\pm0.70}$  &  $\textbf{81.43}_{\pm1.19}$  &  $\textbf{82.77}_{\pm1.13}$  &  $\textbf{80.71}_{\pm3.82}$  &  $\textbf{69.32}_{\pm1.15}$	 & \textbf{82.89}\\
\hline
\end{tabular}}
\end{table}



\subsection{Comparative Study}
We conducted a comparative study to evaluate the performance of our proposed LSCD-TTA method against several state-of-the-art DA and TTA methods using both ResNet-50 and ResNet-101 backbones.
The results are summarized in Tables \ref{tab:methods_50} and \ref{tab:methods_101}.
Methods such as Confidence and ADDA did not perform as well in our experiments. This could be attributed to their reliance on assumptions that may not hold in the complex and diverse domain of RS imagery, where high information density and numerous categories present additional challenges.
\revise{
DA methods such as SHOT, ARS, and CS-CADA, while performing better than other DA methods, may suffer from decreased adaptability due to domain shifts, especially in datasets with high category diversity. These methods aim to minimize the distribution gap between source and target domains, but their performance is still affected by the substantial differences between domains, particularly when the category diversity or environmental variability is high. 
}
TTA methods like Tent, LAME, ARM, DomainAdaptator, \revise{and TRIBE \citep{TRIBE}} series showed relatively better performance but still fell short compared to LSCD-TTA. 
The rapid environmental changes and varying data distributions inherent in RS data can limit these advanced methods’ effectiveness, as they may not adequately account for the unique challenges posed by such datasets.
Notably, Our LSCD-TTA method consistently outperformed all other methods across all tasks. 
Specifically, LSCD-TTA achieved an average accuracy of 81.62\% with ResNet-50 and 82.89\% with ResNet-101, demonstrating significant improvements over the baseline models and other comparative methods.
The method's robust performance across various tasks and backbones underscores its potential for practical applications requiring rapid and accurate adaptation to new and challenging environments.

\revise{
To evaluate the effectiveness and generality of LSCD-TTA compared with other TTA methods, we further extend our experiments to include adaptation tasks with ViT-B/16 \citep{dosovitskiy2020image} as the backbone. As shown in Table \ref{tab:vit}, the results demonstrate that LSCD-TTA exhibits strong applicability and superior performance when applied to transformer-based networks, confirming the generality of our method. However, it is worth noting that the performance of the ViT-based models is slightly lower compared to that of the ResNet-based architectures. CNN architectures tend to perform better than transformer-based networks especially when the dataset size is not extremely large. This is may because CNNs benefit from prior inductive biases, such as local receptive fields and weight sharing, which are particularly advantageous for image tasks. Given that remote sensing images often contain high-dimensional features and rich class information, using CNN-based backbones better utilizes spatial structure information in the absence of massive datasets.
}

\begin{table}[htbp]
\centering
\centering
\caption{Accuracy and standard deviation (\%) of comparative study on RS datasets (Backbone: \textit{ViT-B/16})}
\label{tab:vit}
\renewcommand\arraystretch{1.0}

\resizebox{\linewidth}{!}{
\begin{tabular}{p{2.8cm}<{\centering}|cc|cc|cc|c}
\hline
\multirow{2}{*}{\textbf{Method}} & \multicolumn{2}{|c|}{NWPU-RESISC45} & \multicolumn{2}{|c|}{AID} & \multicolumn{2}{|c|}{UC Merced} & \multirow{2}{*}{\begin{tabular}{c}
    \footnotesize  \textbf{Average}\\
    \footnotesize  \textbf{Accuracy}
\end{tabular}}  \\ 
\cline{2-7}
                         & \textbf{N$\rightarrow$A}      & \textbf{N$\rightarrow$U}     & \textbf{A$\rightarrow$N}        & \textbf{A$\rightarrow$U}        &\textbf{U$\rightarrow$A}        & \textbf{U$\rightarrow$N}              &                        \\
                         \hline
Baseline                & $88.94_{\pm0.31}$  &  $85.60_{\pm0.22}$  &  $76.73_{\pm0.76}$  &  $70.74_{\pm1.69}$  &  $57.72_{\pm1.66}$  &  $56.37_{\pm0.92}$	 & 72.68\\
Confidence              & $88.85_{\pm0.34}$  &  $85.43_{\pm0.24}$  &  $76.43_{\pm0.81}$  &  $70.69_{\pm1.78}$  &  $57.38_{\pm1.66}$  &  $54.94_{\pm1.16}$	 & 72.29\\
CE                      & $89.07_{\pm0.33}$  &  $85.88_{\pm0.28}$  &  $76.88_{\pm0.70}$  &  $70.68_{\pm1.79}$  &  $58.14_{\pm1.72}$  &  $56.56_{\pm1.12}$	 & 72.87\\
KL                      & $88.97_{\pm0.35}$  &  $85.88_{\pm0.28}$  &  $76.86_{\pm0.71}$  &  $70.68_{\pm1.75}$  &  $58.09_{\pm1.68}$  &  $56.64_{\pm1.10}$	 & 72.85\\
CBST                    & $88.99_{\pm0.35}$  &  $85.91_{\pm0.29}$  &  $76.35_{\pm0.76}$  &  $70.66_{\pm1.62}$  &  $58.15_{\pm1.97}$  &  $56.74_{\pm0.99}$	 & 72.80\\
SHOT                    & $89.24_{\pm0.31}$  &  $85.96_{\pm0.38}$  &  $77.56_{\pm0.62}$  &  $71.25_{\pm1.70}$  &  $59.93_{\pm1.80}$  &  $59.29_{\pm0.46}$	 & 73.87\\
ADDA                    & $88.70_{\pm0.39}$  &  $85.57_{\pm0.22}$  &  $75.79_{\pm0.66}$  &  $70.52_{\pm1.68}$  &  $57.56_{\pm1.47}$  &  $56.25_{\pm0.38}$	 & 72.40\\
ARM                     & $88.96_{\pm0.35}$  &  $85.52_{\pm0.71}$  &  $76.90_{\pm0.53}$  &  $70.85_{\pm0.89}$  &  $58.67_{\pm1.44}$  &  $55.91_{\pm1.66}$	 & 72.80\\
Tent                    & $89.15_{\pm0.33}$  &  $85.96_{\pm0.33}$  &  $76.88_{\pm0.77}$  &  $70.75_{\pm1.82}$  &  $58.15_{\pm1.79}$  &  $56.32_{\pm1.25}$	 & 72.87\\
SLR                     & $89.24_{\pm0.49}$  &  $85.55_{\pm0.53}$  &  $73.98_{\pm0.99}$  &  $69.20_{\pm1.69}$  &  $43.17_{\pm3.06}$  &  $52.62_{\pm1.25}$	 & 68.96\\
\hline
\textbf{LSCD-TTA}      & $\textbf{90.40}_{\pm0.36}$  &  $\textbf{87.54}_{\pm0.17}$  &  $\textbf{78.78}_{\pm0.78}$  &  $\textbf{73.23}_{\pm2.20}$  &  $\textbf{65.22}_{\pm2.54}$  &  $\textbf{62.27}_{\pm1.21}$	 & \textbf{76.24}\\
\hline
\end{tabular}
}
\end{table}

\revise{
To objectively assess time efficiency, we analyze and compare the adaptation time per image, calculating the average execution time over multiple experimental runs. As shown in Table \ref{tab:time}, the baseline method, which does not perform any adaptation, represents the lower bound of time consumption at 0.70 ms/image. However, this comes at the cost of poor adaptation performance. In contrast, despite the added complexity from the novel loss functions, LSCD-TTA maintains an average time consumption of 2.65 ms/img, outperforming all existing TTA methods. This demonstrates that LSCD-TTA remains within acceptable time limits and is a practical solution for real-world applications. Therefore, LSCD-TTA’s time efficiency and accuracy make it particularly suitable for real-time remote sensing applications where quick adaptation is essential.
}

\begin{table}[htbp]
\centering
\caption{Time Consumption Comparison of TTA methods with \textit{Resnet-50}}
\label{tab:time}
\renewcommand\arraystretch{1.0}
\fontsize{8}{10}\selectfont
\resizebox{\linewidth}{!}{
\revise{
\begin{tabular}{p{1.8cm}<{\centering}|cc|cc|cc|c}
\hline
\multirow{2}{*}{\textbf{Method}} & \multicolumn{2}{c|}{NWPU-RESISC45} & \multicolumn{2}{c|}{AID} & \multicolumn{2}{c|}{UC Merced} & \multirow{2}{*}{\begin{tabular}{c}
    \fontsize{8}{10}\selectfont\textbf{Average} \\
    \textbf{\textit{(ms/img)}}
\end{tabular}}  \\ 
\cline{2-7}
                         & \textbf{N$\rightarrow$A}      & \textbf{N$\rightarrow$U}     & \textbf{A$\rightarrow$N}        & \textbf{A$\rightarrow$U}        &\textbf{U$\rightarrow$A}        & \textbf{U$\rightarrow$N}              &                        \\
                         \hline
Baseline                 & 0.62  & 0.58  & 0.64  & 0.89  & 0.86  & 0.63  & 0.70 \\
\hline
LAME                     & 2.98  & 5.38  & 3.03  & 4.97  & 7.38  & 4.05  & 4.63 \\
Tent                     & 2.65  & 6.27  & \textbf{2.37}  & 3.26  & 4.43  & 5.99  & 4.16 \\
SLR                      & 8.11  & 12.07 & 8.21  & 10.26 & 11.34 & 8.96  & 9.83 \\
DA-T                     & 8.26  & 12.09 & 9.44  & 4.79  & 5.05  & 10.66 & 8.38 \\
DA-SKD                   & 12.64 & 15.86 & 12.47 & 16.08 & 16.22 & 11.99 & 14.21 \\
DA-AUG                   & 14.77 & 8.73  & 15.58 & 6.74  & 6.97  & 7.33  & 10.02 \\
{TRIBE} & {8.59}  & {8.57}  & {8.42}  & {8.57}  & {8.64}  & {8.97}  & {8.63} \\
\hline
\textbf{LSCD-TTA}       & \textbf{2.41}  & \textbf{2.51}  & 2.39  & \textbf{3.17}  & \textbf{3.44}  & \textbf{1.97}  & \textbf{2.65} \\
\hline
\end{tabular}}
}
\end{table}

\subsection{Ablation study}
As shown in Tab. \ref{tab:ablation}, the ablation experiments employ the average accuracy to assess the effectiveness of the three proposed losses. 
The source-trained model achieves an average accuracy of 74.19 \% and 75.19 \% for Resnet-50 and Resnet-101. 
In this experiment, the proposed WCSE, BCSE, and LSD losses provide individual improvements in average accuracy of 4.04\%, 3.17\%, and 6.45\% for ResNet-50, and 4.38\%, 3.53\%, and 6.82\% for ResNet-101.
Combining $\mathcal{L}_{wcse}$ with $\mathcal{L}_{wcse}$ results in a performance boost, achieving an average accuracy of 78.16\% and 79.49\% for Resnet-50 and Resnet-101.
This suggests that addressing both the uncertainty and category diversity simultaneously provides joint benefits, enhancing the model’s ability to generalize across diverse target domain distributions.

On the other hand, incorporating the individual loss $\mathcal{L}_{lsd}$ with $\mathcal{L}_{wcse}$ and $\mathcal{L}_{bcse}$ significantly raises the performance to 81.34\% and 81.19\% for ResNet-50, and 82.74\% and 82.18\% for ResNet-101. 
This suggests that while increasing attention to weak categories or diversity can yield better results, there may be potential redundancy or conflicting optimization objectives between these two loss functions.
Additionally, excessive preference may lead to underestimating or even misclassifying predictions that should have been trusted, causing a performance bottleneck. After incorporating $\mathcal{L}_{lsd}$, we apply the desaturation concept to make the model more cautious in its decision-making, further enhancing the cross-domain adaptation performance and the robustness of the overall system in real time.

Remarkably, our proposed LSCD-TTA method, which integrates all three loss functions, achieves the best performance with an average accuracy of 81.62\% and 82.89\%. 
This indicates that the combined effect of three losses synergistically enhances the model’s adaptation capability.

\begin{table}[htbp]
    \centering
    \setlength{\tabcolsep}{3mm}{
    \renewcommand\arraystretch{1.1}
    \caption{Ablation study of the LSCD-TTA}
    \label{tab:ablation}
    \resizebox{0.9\linewidth}{!}{
    \begin{threeparttable}
    \begin{tabular}{c|ccc|cccccc|c}
\hline
\multicolumn{11}{c}{\textit{Resnet-50} \citep{resnet} } \\
\hline
Method & $\mathcal{L}_{wcse}$ & $\mathcal{L}_{bcse}$ & $\mathcal{L}_{lsd}$ & N$\rightarrow$A & N$\rightarrow$U & A$\rightarrow$N & A$\rightarrow$U & U$\rightarrow$A & U$\rightarrow$N & Average Accuracy   \\ \hline
Baseline    &     &     &       & 89.74  & 83.98  & 78.87  & 72.49 & 62.11  & 58.04   & 74.19  \\ 
\hline
(A)    & \Checkmark    &     &       & 91.74  &  88.18  &  79.54  &  79.54  &  70.27  &  60.13	 & 78.23  \\
(B)    &     &\Checkmark     &    & 90.45  &  86.17  &  80.17  &  76.49  &  67.47  &  63.40	 & 77.36     \\
(C)    &     &     &\Checkmark       & 91.61  &  88.16  &  \textbf{81.99}  &  79.15  &  74.09  &  68.81	 & 80.64  \\
\hline
(D)    & \Checkmark    & \Checkmark    &       & 91.29                & 86.84               & 80.52               & 77.85               & 69.64               & 62.82               & 78.16  \\
(E)    & \Checkmark    &     &\Checkmark       & 92.15                & 88.27               & 81.13               & 81.22               & \textbf{77.20}               & 68.05              & 81.34  \\
(F)    &     & \Checkmark    &\Checkmark       & 92.07                & 88.24               & 80.92              & 81.08               & 76.24               & 68.57               & 81.19  \\
\hline
(G)    & \Checkmark    & \Checkmark    & \Checkmark     & \textbf{92.32}               & \textbf{88.73}              & 81.70              & \textbf{81.46}               & 76.36               & \textbf{69.07}               & \textbf{81.62}  \\ 
\hline
\multicolumn{11}{c}{}\\
\hline
\multicolumn{11}{c}{\textit{Resnet-101} \citep{resnet} } \\
\hline
Method & $\mathcal{L}_{wcse}$ & $\mathcal{L}_{bcse}$ & $\mathcal{L}_{lsd}$ & N$\rightarrow$A & N$\rightarrow$U & A$\rightarrow$N & A$\rightarrow$U & U$\rightarrow$A & U$\rightarrow$N & Average Accuracy   \\ \hline
Baseline  &     & &       & 89.98  & 87.26  & 80.08  & 71.81  & 63.42 & 58.60   & 75.19   \\
\hline
(A)    & \Checkmark    &     &       & 92.08  &  91.06  &  81.12  &  81.40  &  73.57  &  58.22	 & 79.57\\
(B)    &     & \Checkmark    &       & 91.18  &  88.43  &  81.49  &  77.68  &  69.92  &  63.62	 & 78.72 \\
(C)    &     &     &  \Checkmark     & 92.23  &  89.59  &  \textbf{82.91}  &  80.26  &  77.03  &  \textbf{70.04} & 82.01  \\
\hline
(D)    & \Checkmark    & \Checkmark    &       & 91.65         & 89.87               & 81.74              & 79.55               & 72.19               & 61.94               & 79.49  \\
(E)    & \Checkmark    &     &\Checkmark       & 92.19         & 90.40               & 81.68               & 82.60               & 80.49               & 69.10             & 82.74  \\
(F)    &     & \Checkmark    &\Checkmark       & 92.18         & 90.14               & 81.07              & 82.18               & 80.19               & 67.30               & 82.18  \\
\hline
(G)    & \Checkmark    & \Checkmark    & \Checkmark     & \textbf{92.43}       & \textbf{90.65}        & 81.43           & \textbf{82.77}             & \textbf{80.71}               &69.32                & \textbf{82.89}  \\ 
\hline
\end{tabular}
\end{threeparttable}
}
}
\vspace{0.3em}
\end{table}

\begin{figure}[htbp]
    \centering
    \includegraphics[width=.90\linewidth]{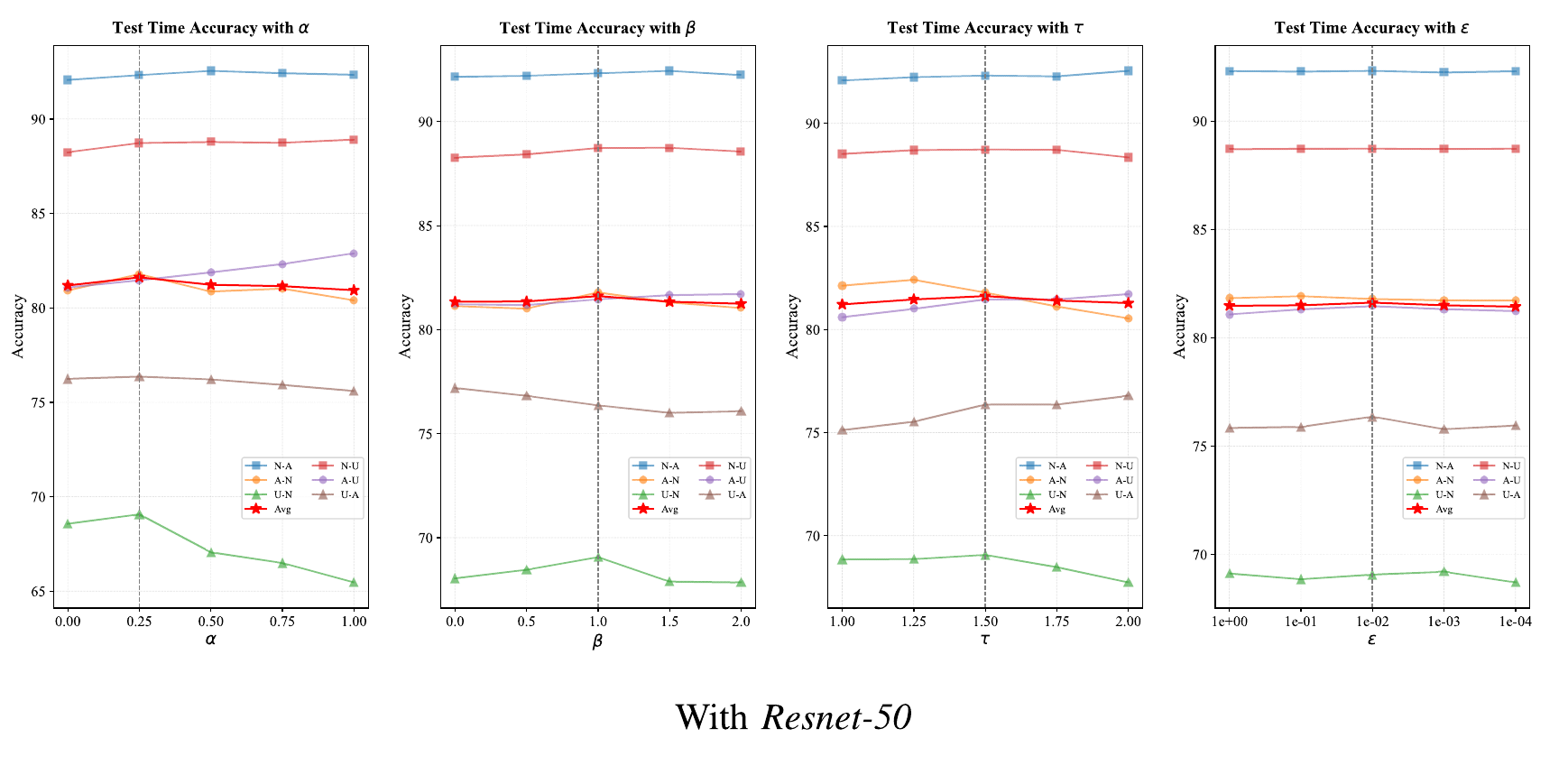}
    \vspace{-5pt}
    \includegraphics[width=.90\linewidth]{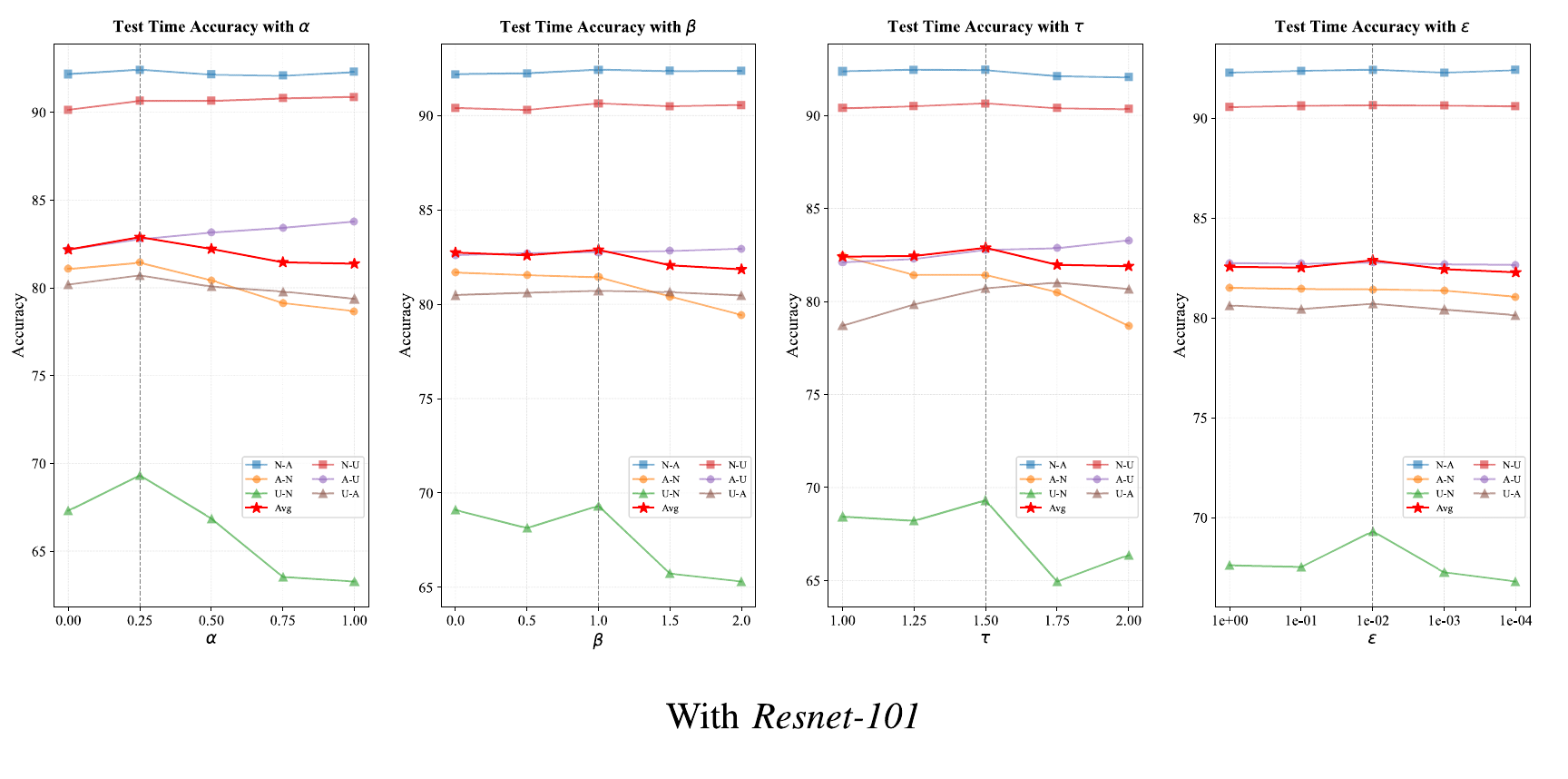}
    \caption{Sensitivity analysis of $\alpha$, $\beta$, $\tau$ with \textit{Resnet-50} (top) and \textit{Resnet-101} (bottom)}
    \label{fig:ablation_50}
\end{figure}

\subsection{Sensitivity analysis of the hyperparameters $\alpha,\beta,\tau$}
As illustrated in Fig. \ref{fig:ablation_50}, we further conduct sensitivity studies on the hyperparameters $\alpha,\beta,\tau \revise{,\epsilon}$ with different backbones separately to find the relatively best setting of the hyperparameter used in LSCD-TTA.
Our findings indicate that different hyperparameter configurations can yield superior performance in specific DA tasks. 
However, the overall performance trends remained relatively consistent, suggesting that LSCD-TTA maintains robust effectiveness across a range of hyperparameter values.
Notably, we \revise{identify} the optimal combined performance with $\alpha = 0.25, \beta = 1, \tau = 1.5, \revise{\epsilon = 0.01}$ when using ResNet-50 and ResNet-101. 
These settings appear to balance the contributions of each loss component effectively.
 
\begin{figure}[htbp]
  \centering
  \captionsetup[subfigure]{labelformat=empty} 
  \begin{subfigure}
    \centering
    \includegraphics[width=.41\linewidth]{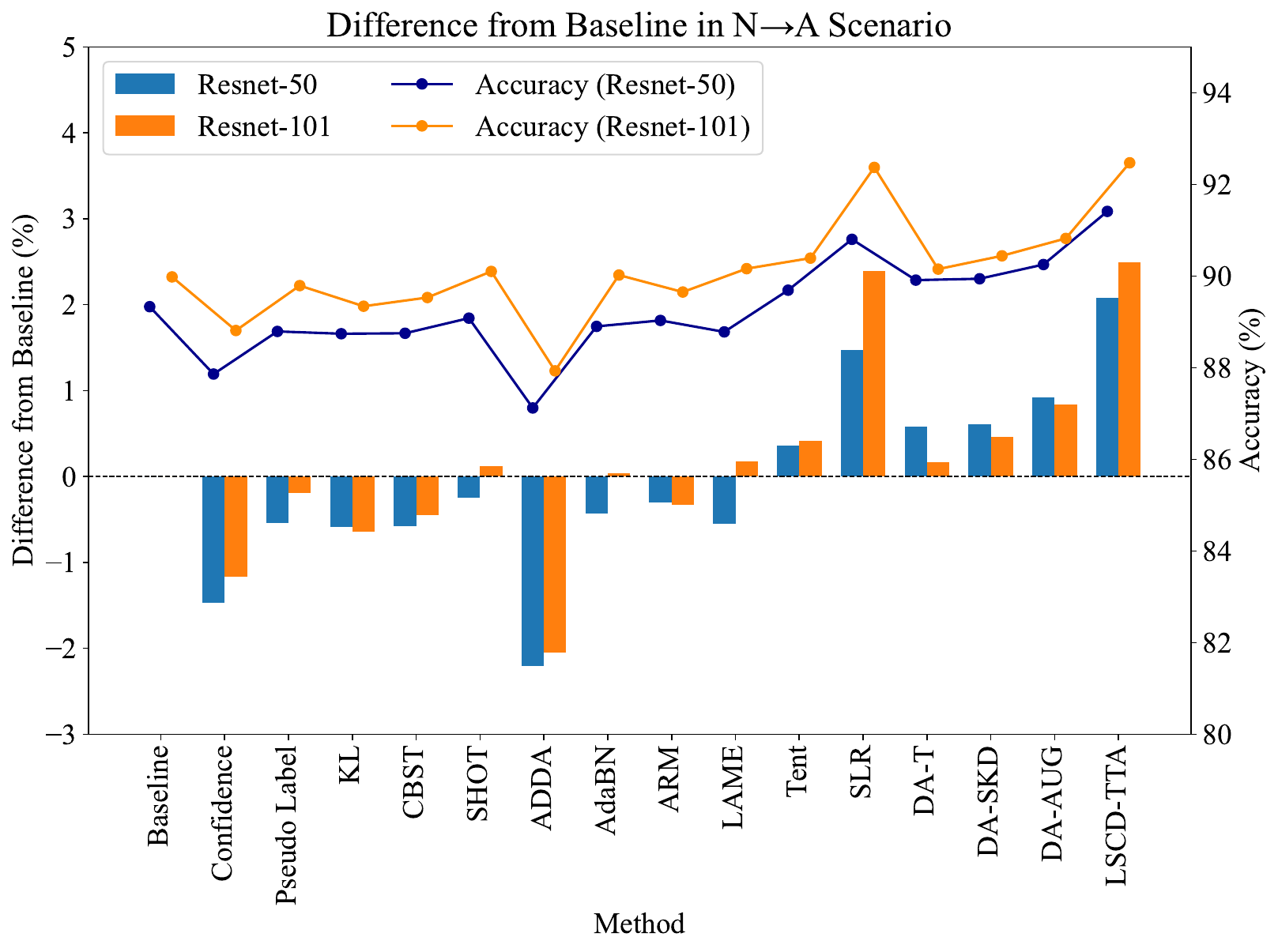}
    \includegraphics[width=.41\linewidth]{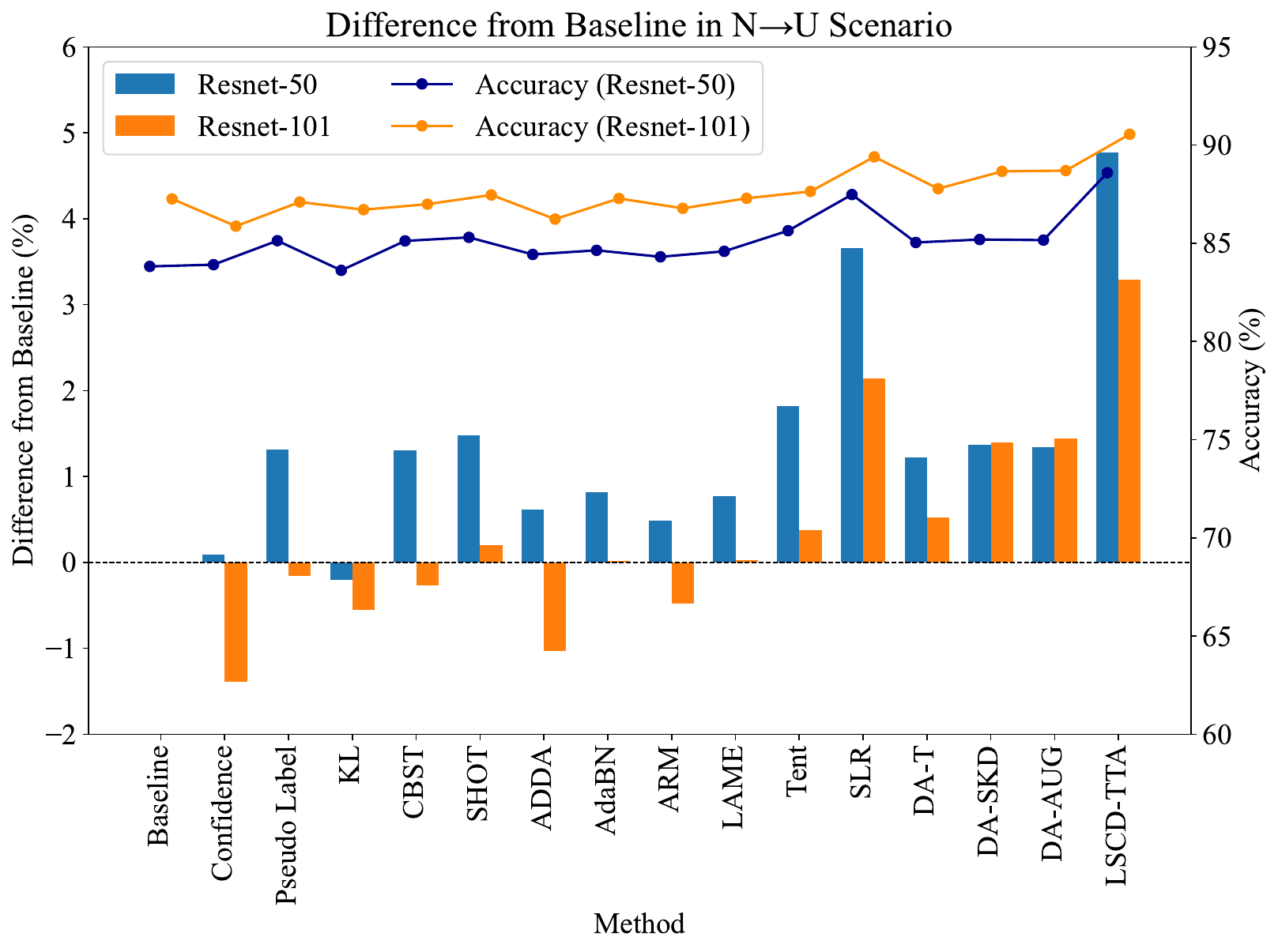}
    \\
    \includegraphics[width=.41\linewidth]{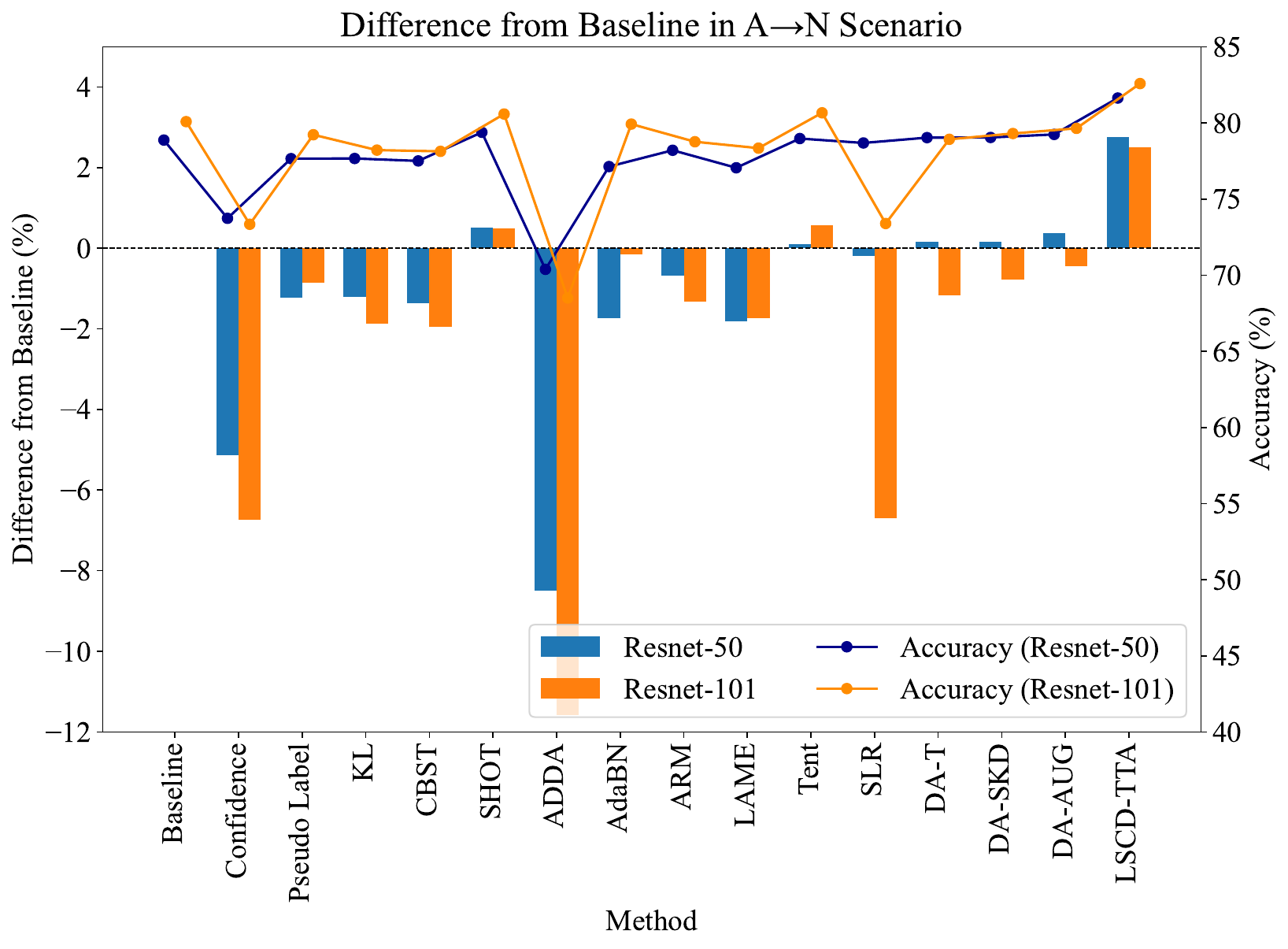}
    \includegraphics[width=.41\linewidth]{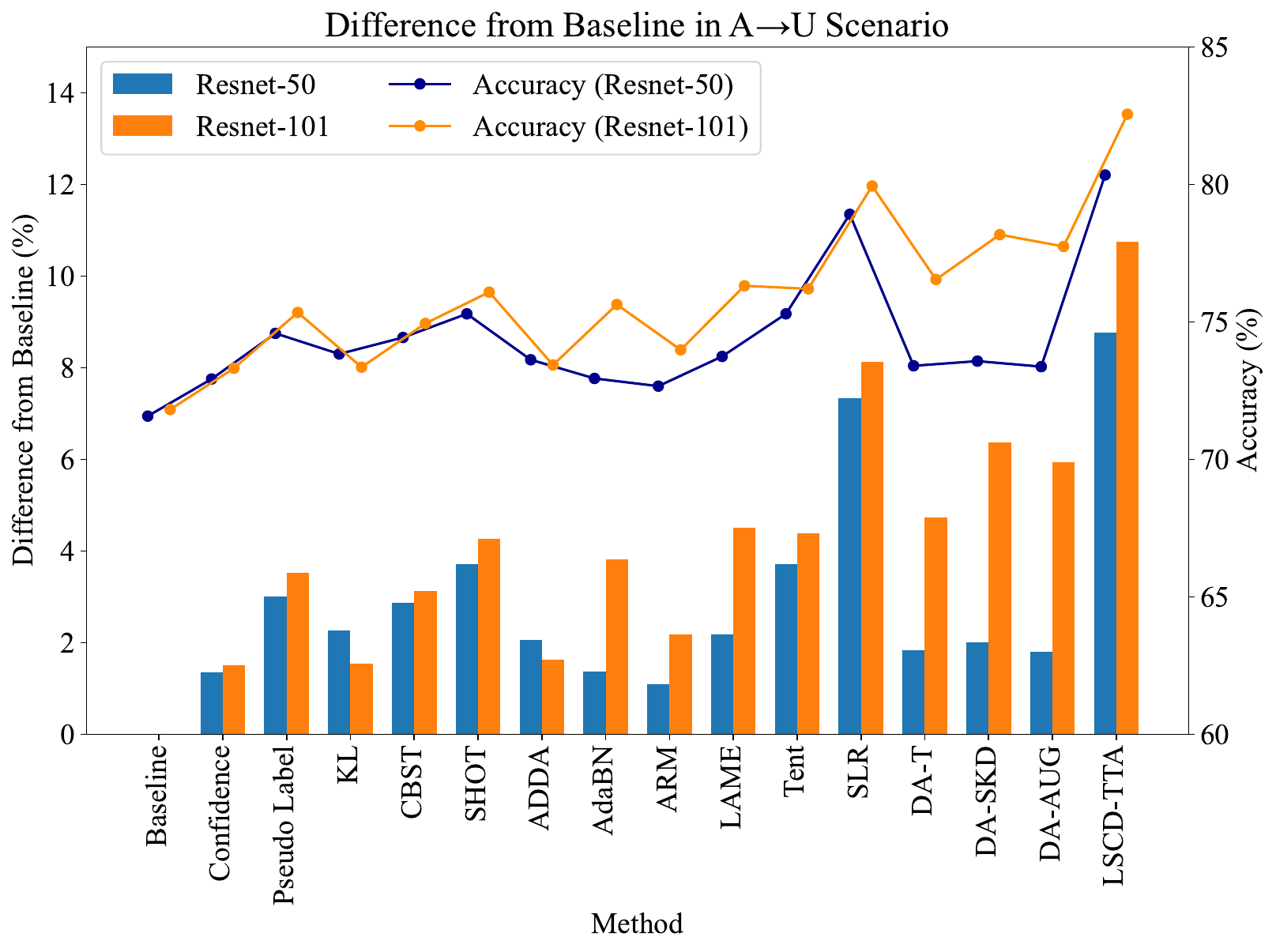}
    \\
    \includegraphics[width=.41\linewidth]{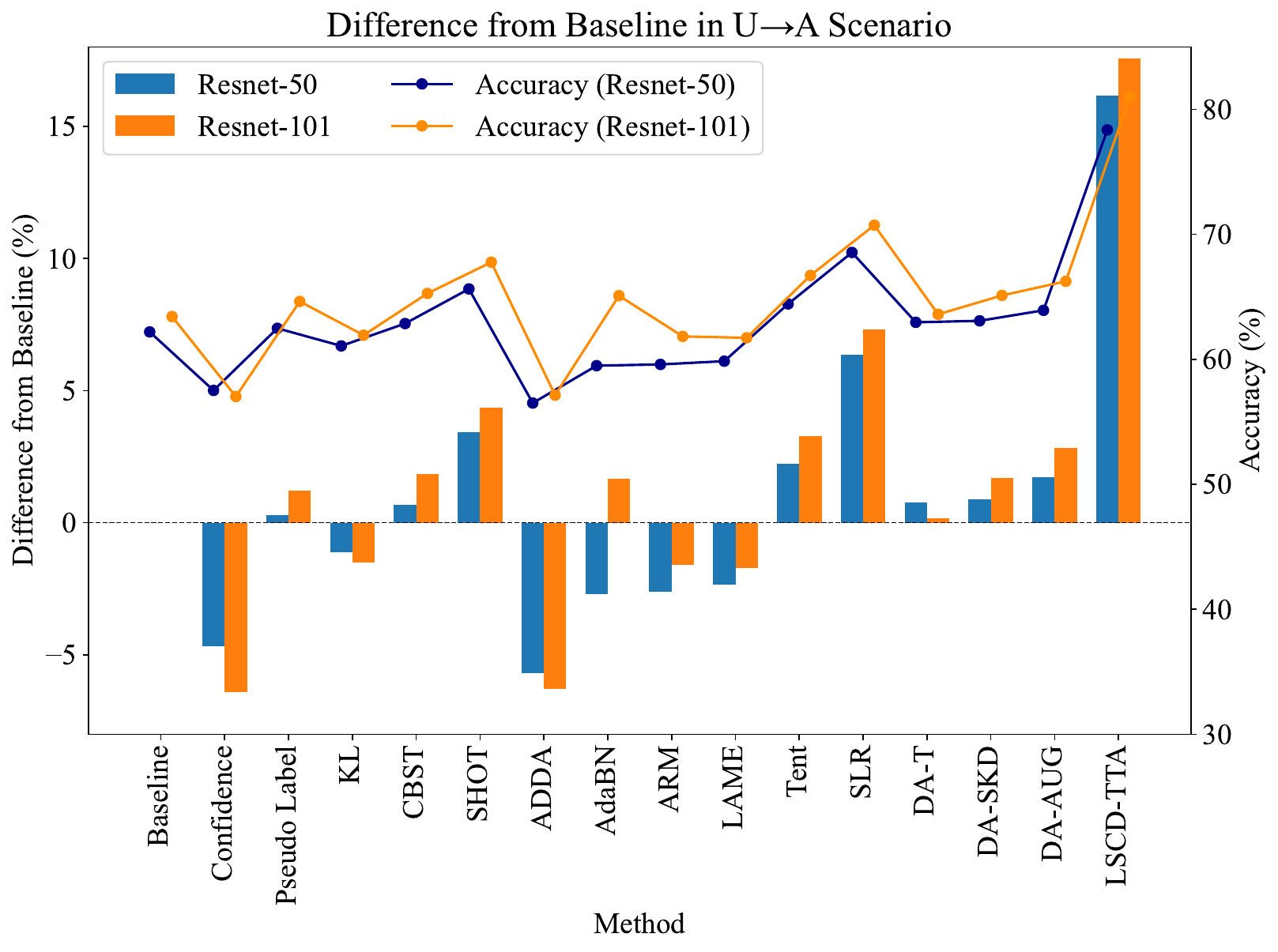}
    \includegraphics[width=.41\linewidth]{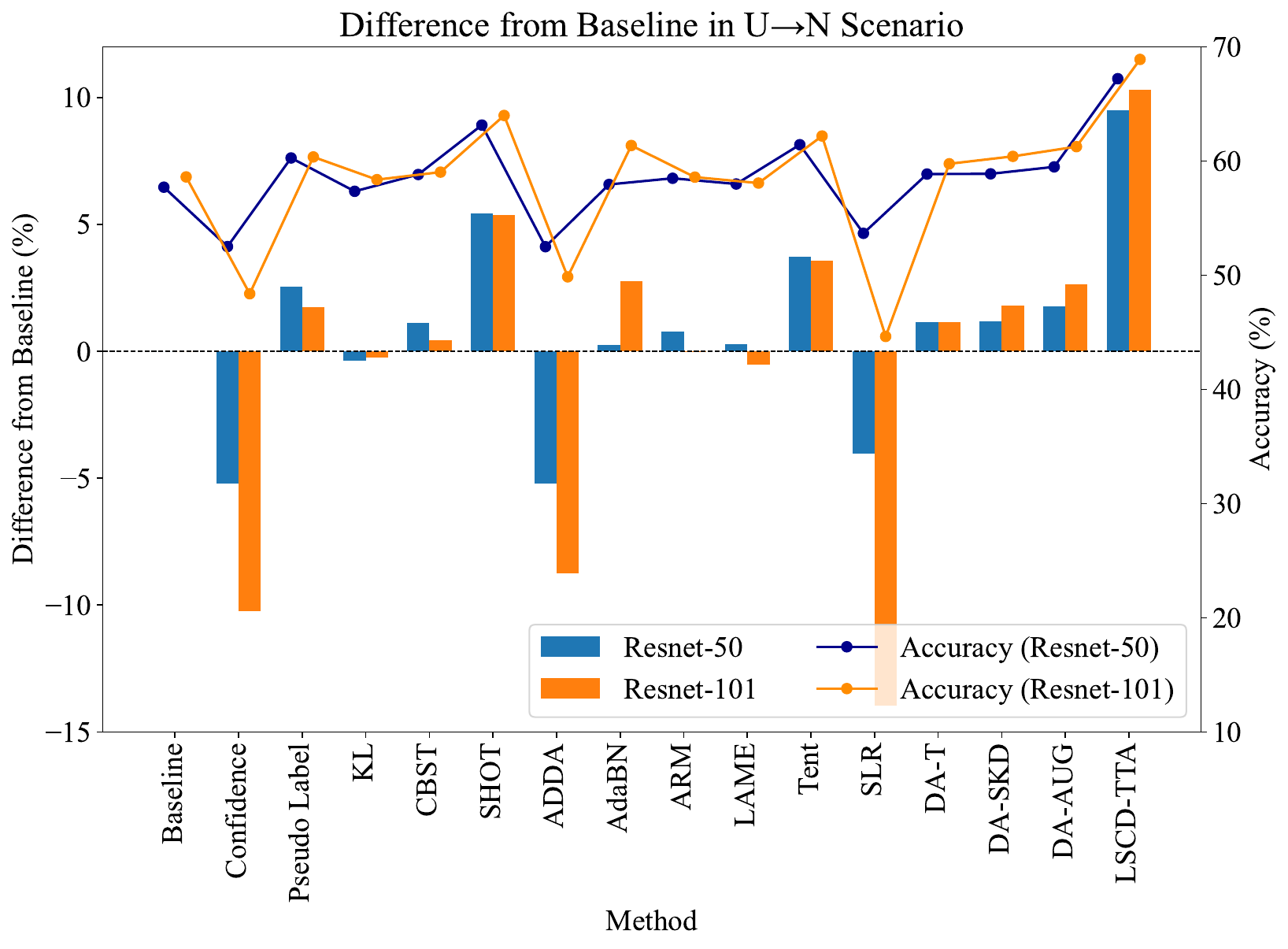}
    \caption{
    Negative adaptation results with \textit{Resnet-50} and \textit{Resnet-101}.
    Compared to other DA and TTA methods, our proposed LSCD-TTA demonstrates superior performances in adaptation tasks in terms of average accuracy.
    }
    \label{fig:negative}
  \end{subfigure}
\end{figure}

\section{Discussion}
\label{sec:dis}
\vspace{0.4em}

\subsection{Negative transfer}
To provide an intuitive comparison of different methods against the baseline, we plot a negative transfer graph (see Fig. \ref{fig:negative}) to visualize the overall performance relative to the baseline.
Results demonstrate that LSCD-TTA consistently demonstrates state-of-the-art performance across nearly all DA tasks.
Notably, in the U$\rightarrow$A and U$\rightarrow$N tasks, where the source domain models are derived from a limited knowledge domain and subsequently adapt to information-rich target domains, LSCD-TTA exhibits particularly outstanding results. 
This indicates that our method effectively enables models to efficiently and robustly adapt to broader and more complex target domains.

In contrast, while SLR \citep{slr} shows competitive performance in the N$\rightarrow$A, N$\rightarrow$U, and A$\rightarrow$U tasks, it suffers from negative transfer effects in the A$\rightarrow$N and U$\rightarrow$N tasks. 
Compared to these fluctuating results, LSCD-TTA produces more stable and consistent outcomes across all tasks, highlighting its robustness.
These observations verify that LSCD-TTA is well-suited for RS datasets characterized by numerous categories, high sample similarity, and frequent data changes. 
By effectively enhancing performance in cross-domain classification tasks, LSCD-TTA addresses the challenges posed by negative transfer, ensuring reliable adaptation in complex RS applications.

\subsection{Visualization}
\begin{figure}[htbp]
    \centering
    \includegraphics[width=\linewidth]{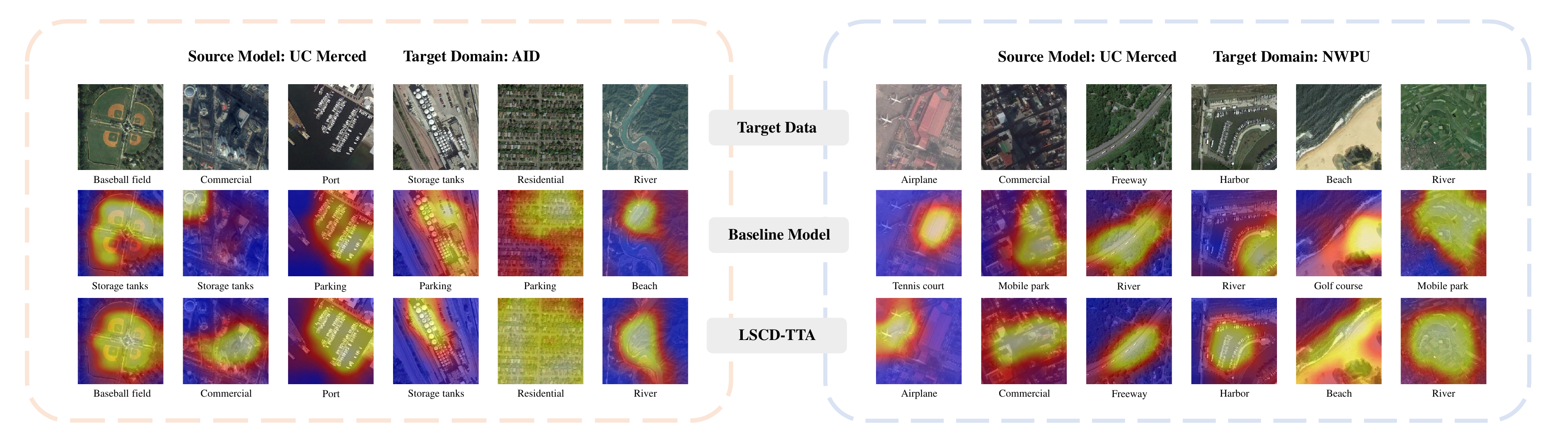}
    \caption{Grad-CAM comparison between LSCD-TTA and the baseline model using \textit{ResNet-50}. The heatmaps illustrate the areas of focus during different adaptation tasks. The baseline model’s attention is constrained by limited source knowledge, often leading to incorrect focus (e.g., misidentifying commercial buildings as storage tanks in U$\rightarrow$A) or misclassification despite correct attention areas (e.g., misclassifying ports as parking lots in U$\rightarrow$A). 
    In contrast, LSCD-TTA enhances attention performance and enables fine-grained classification while preserving global semantic information.
    }
    \label{fig:gradcam}
\end{figure}

To demonstrate the effectiveness of LSCD-TTA, we employ Grad-CAM \citep{gradcam} to visualize the model’s attention mechanisms. 
Grad-CAM generates class-specific attention maps by computing the gradients of class scores with respect to convolutional feature maps. 
As shown in Figure \ref{fig:gradcam}, the baseline model, which is trained with limited source domain knowledge, focuses on image regions that resemble source domain priors (e.g., rivers and freeways, parking lots and ports), which can lead to incorrect predictions.
In contrast, LSCD-TTA enables the model to adaptively adjust its focus during test time, paying attention to both prominent features and harder-to-classify regions. 
This flexibility allows the model to balance global and local feature considerations, resulting in more accurate and robust classification in the target domain.

\subsection{Future outlook}
While we address DA through feature-aligned losses and domain offset minimization, there are instances where feature distributions are not well-aligned across broader areas, potentially degrading performance, especially with large domain shifts.
For future work, we plan to explore more effective feature alignment methods and advanced DA strategies to enhance the model’s adaptability and generalization capabilities. 
Additionally, our current approach assumes a one-to-one category correspondence between source and target domains.
However, in real-world scenarios, models often encounter entirely new categories due to diverse geographic environments. 
Therefore, we aim to investigate methods that improve the model’s ability to generalize to unknown domains and novel classes, further extending the applicability of LSCD-TTA in RS tasks.

\section{Conclusion}
\label{sec:conc}
In this paper, We propose LSCD-TTA, a novel test-time adaptation method tailored for cross-scene RS image classification. 
LSCD-TTA introduces three specialized loss functions: the WCSE loss improves the model’s ability to distinguish difficult-to-classify categories; the BCSE loss mitigates deviations in cross-category sample distributions by softening and balancing predicted probability distributions; and the LSD loss encourages the model to be more cautious with each prediction, allowing it to find the correct adaptation direction through stable steps, thereby improving performance in the later stages of adaptation.
Extensive experiments conducted on three RS datasets (\textit{i.e.}, NWPU-RESISC45, AID, and UC Merced) demonstrate that LSCD-TTA significantly outperforms state-of-the-art DA and TTA methods. Specifically, it achieves average accuracy improvements of 4.99$\%$ with ResNet-50, 5.22$\%$ with ResNet-101, \revise{and 2.37$\%$ with ViT-B/16}.
These results confirm that LSCD-TTA effectively addresses the challenges of cross-scene classification in RS imagery, enabling source models to generalize quickly and accurately to diverse target domains.
We anticipate that LSCD-TTA will have broad applications in practical and generalized DA scenarios within the RS community.
\section*{Acknowledgments}
This work was supported by the National Natural Science Foundation of China (Grant No. T2125006 and 42401415) and Jiangsu Innovation Capacity Building Program (Project No. BM2022028).
\bibliographystyle{elsarticle-harv} 
\bibliography{cas-refs}





\end{document}